\documentclass[10pt,twocolumn,letterpaper]{article}

\usepackage[preprint, pagenumbers]{cvpr} % To force page numbers, e.g. for an arXiv version

\definecolor{cvprblue}{rgb}{0.21,0.49,0.74}
\usepackage[pagebackref,breaklinks,colorlinks,allcolors=cvprblue]{hyperref}
\usepackage{multirow}
\usepackage{tcolorbox}

\def\paperID{11402} % *** Enter the Paper ID here
\def\confName{CVPR}
\def\confYear{2025}

\newcommand\ntfootnote[1]{%
  \begin{NoHyper}
  \renewcommand\thefootnote{}\footnotetext{#1}%
  \addtocounter{footnote}{0}%
  \end{NoHyper}
}

\title{Auto Cherry-Picker \raisebox{-0.0001\height}{\includegraphics[height=0.6cm]{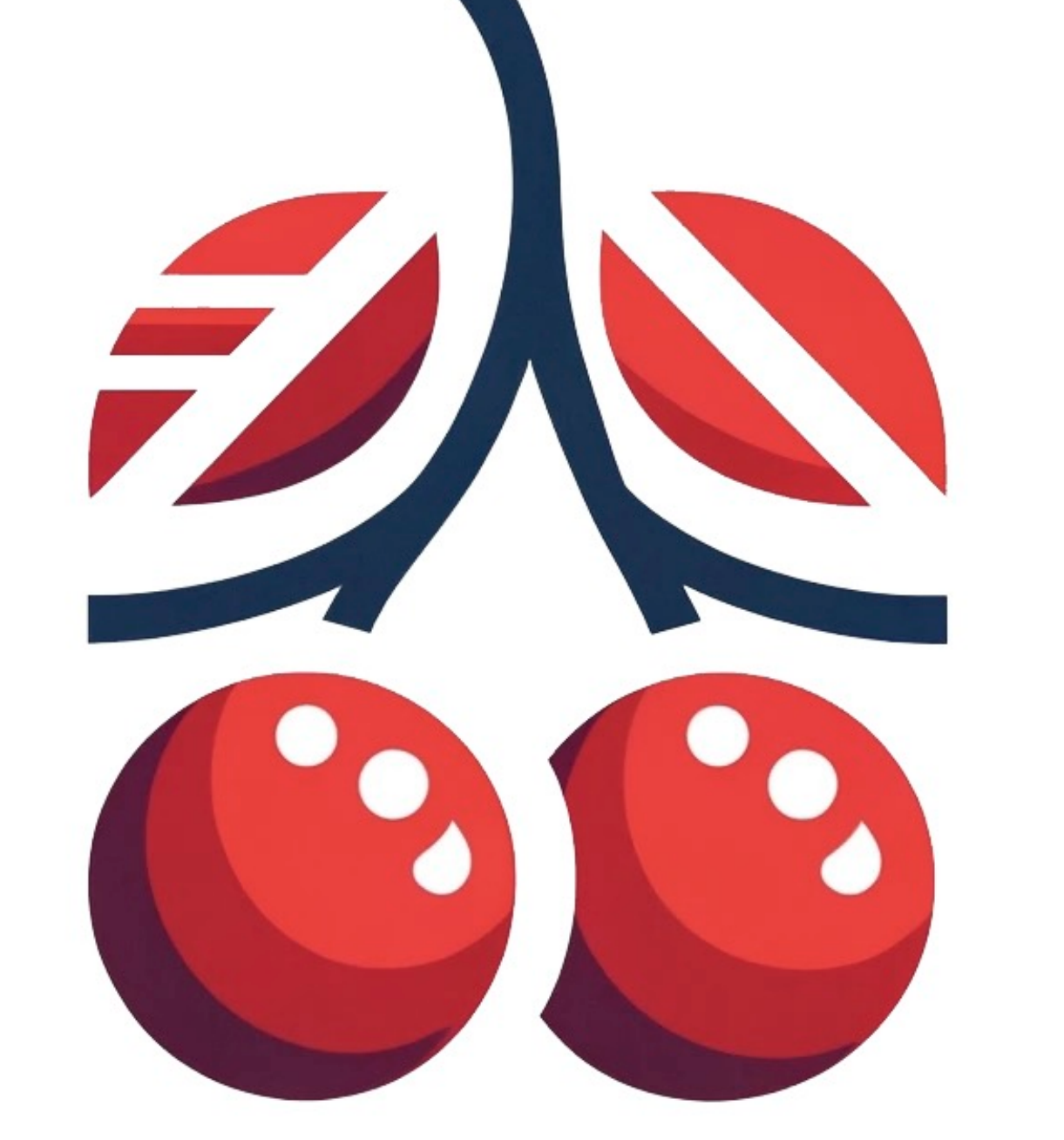}}: Learning from High-quality \\ Generative Data Driven by Language}

\author{\textbf{Yicheng Chen$^{1,2}$, Xiangtai Li$^{1,3}$, Yining Li$^{1\dag}$, Yanhong Zeng$^{1}$,}\\
\textbf{Jianzong Wu$^{1,4}$, Xiangyu Zhao$^{1,5}$, Kai Chen$^{1\dag}$} \\
  {$^{1}$Shanghai AI Laboratory}
  {$^{2}$Fudan University} \\
  {$^{3}$S-Lab, Nanyang Technological University} \\
  {$^{4}$Peking University}
  {$^{5}$Shanghai Jiao Tong University} \\
  \textbf{Project page}: \href{https://yichengchen24.github.io/projects/autocherrypicker}{https://yichengchen24.github.io/projects/autocherrypicker}
}

\begin{document}
% \maketitle
\twocolumn[{%
  \renewcommand\twocolumn[1][]{#1}%
  \vspace{-56pt}
  \maketitle
    \vspace{-12pt}
    \captionsetup{type=figure}
    \centering
    \includegraphics[width=.9\textwidth]{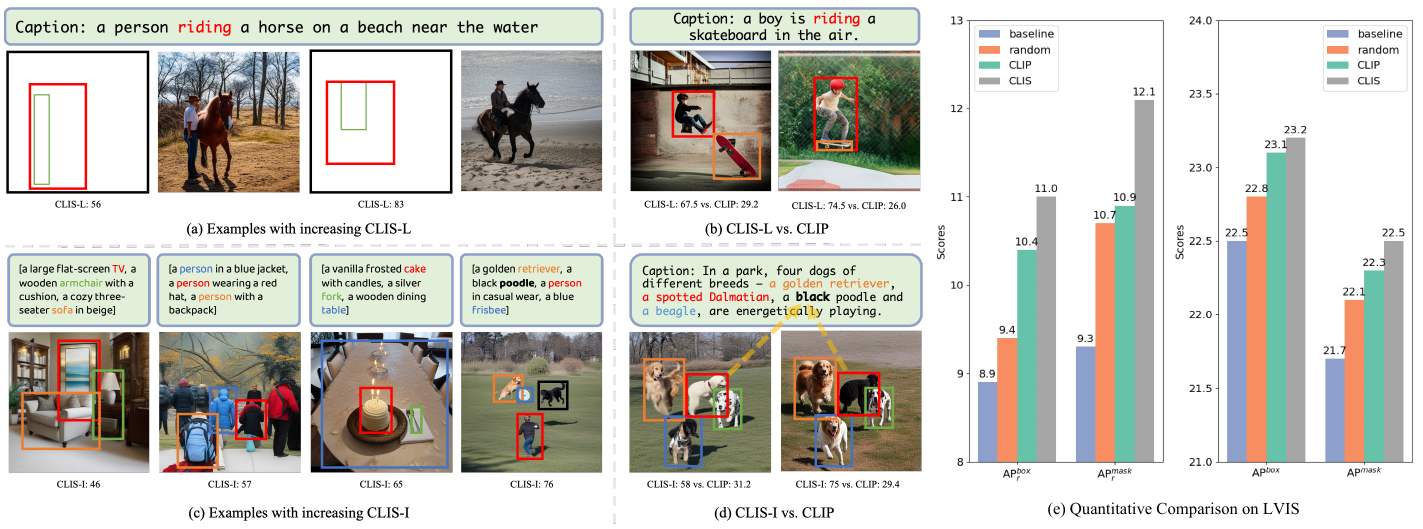}
    \vspace{-8pt}
    \caption{Illustration of quality assessment of generated data samples using CLIS. (a) and (c) compare the quality of samples with different CLIS-L and CLIS-I scores, respectively. Samples with low CLIS fail to align accurately with the condition (e.g., contain extraneous objects or exhibit visual flaws). (b) and (d) compare the preferences of CLIS and CLIP score~\cite{hessel2021clipscore}. (e) compares different selection methods for the same volume of synthetic data used in downstream tasks, reporting AP$_r$ and AP on the LVIS benchmark. See details in Sec.~\ref{sec:stu-eff}.}
    \label{fig:teaser}
    \vspace{12pt}
}]
\ntfootnote{$^{\dag}$ Corresponding Author.}
\begin{abstract}
Diffusion models can generate realistic and diverse images, potentially facilitating data availability for data-intensive perception tasks.
However, leveraging these models to boost performance on downstream tasks with synthetic data poses several challenges, including aligning with real data distribution, scaling synthetic sample volumes, and ensuring their quality. 
To bridge these gaps, we present \textbf{A}uto \textbf{C}herry-\textbf{P}icker (ACP), a novel framework that generates high-quality cross-modality training samples at scale 
to augment perception and multi-modal training. 
ACP first uses LLMs to sample descriptions and layouts based on object combinations from real data priors, eliminating the need for ground truth image captions or annotations. 
Next, we use an off-the-shelf controllable diffusion model to generate multiple images. 
Then, the generated data are refined using a comprehensively designed metric, Composite Layout and Image Score (CLIS), to ensure quality.
Our customized synthetic high-quality samples boost performance in various scenarios, especially in addressing challenges associated with long-tailed distribution and imbalanced datasets.
Experiment results on downstream tasks demonstrate that ACP can significantly improve the performance of existing models. 
In addition, we find a positive correlation between CLIS and performance gains in downstream tasks.
This finding shows the potential for evaluation metrics as the role for various visual perception and MLLM tasks.
\end{abstract}
    
\section{Introduction}
\label{sec:intro}

Recently, diffusion-based image generation methods~\cite{rombach2022high,sinha2021d2c} have made remarkable progress, enabling various applications, including text-to-image generation(T2I)~\cite{han2023generalist,han2024face,ramesh2022hierarchical,saharia2022photorealistic,wang2024semflow}, image editing~\cite{brooks2023instructpix2pix,hertz2022prompt,meng2021sdedit,saharia2022palette}, video generation~\cite{harvey2022flexible,ho2022imagen,ho2022video}, and more. 
Compared to previous generative models~\cite{esser2021taming,isola2017image}, diffusion models can produce \textbf{high-quality} and \textbf{high-resolution} examples.
Thus, one essential usage of the diffusion-based model is to generate training data for various downstream vision tasks, such as segmentation~\cite{li2023open,xie2023mosaicfusion}, detection~\cite{chen2023geodiffusion,lin2023explore}, and visual representation learning~\cite{li2023dreamteacher,tian2024stablerep}. 
Synthetic data alleviates the severe demand for human annotation and provides a more controllable data production process.

\begin{figure}[t]
    \centering
    \includegraphics[width=.9\linewidth]{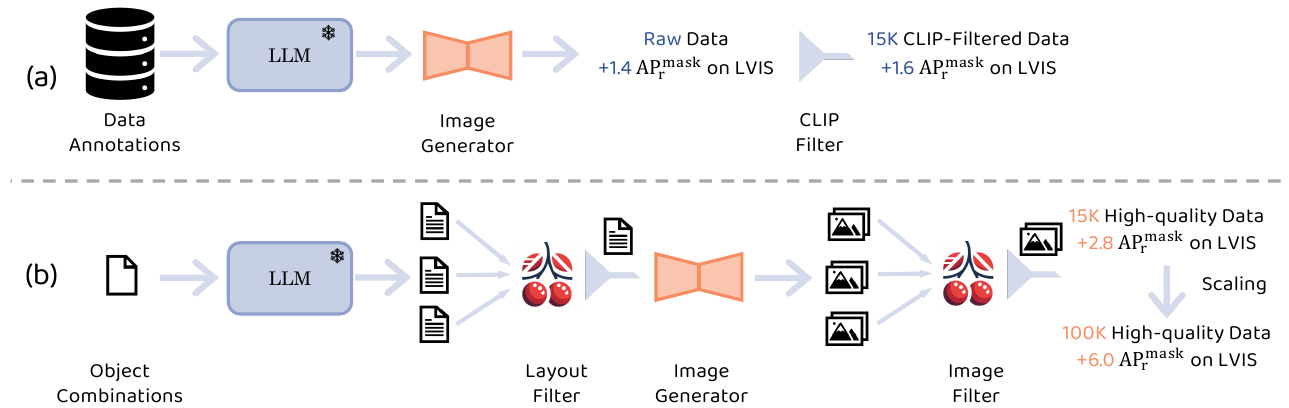}
    \caption{\footnotesize Comparison with previous methods. (a) LMD~\cite{lian2024llmgrounded} generates samples conditioned on detailed image descriptions by leveraging LLMs as the layout generator and diffusion-based models as the image generator. Some methods~\cite{he2022synthetic} use CLIP filtering to future refine these samples. (b) ACP synthesizes training samples conditioned solely on object combinations in natural language and automatically cherry-picks high-quality ones by evaluating both layouts and images. High-quality training samples are more effective for downstream tasks.}
    \vspace{-12pt}
    \label{fig:method-comp}
\end{figure}

Previous works have explored generating samples via different types of references, such as captions~\cite{lian2024llmgrounded}, bounding boxes~\cite{wang2024instancediffusion}, instance masks~\cite{liu2019pixel,tan2023diffss}, and reference images~\cite{ye2023seggen}.
In particular, InstanceDiffusion~\cite{wang2024instancediffusion} 
generates images with precise instance-level control, including attribute binding and localization, while maintaining high fidelity conditioned on detailed instances.
However, these approaches still rely on \textit{manual} annotations, limiting their scalability and diversity in generating large-scale training datasets for downstream tasks.
In addition, due to the \textit{inherent randomness} of generative models, the quality of generated data tends to vary, potentially impairing the effectiveness of using this data to train downstream tasks~\cite{he2022synthetic}. 
Consequently, appropriate quality assessment metrics must be employed to rule out low-quality synthetic samples. 
At the image level, current semantic-based metrics, such as CLIP-Score~\cite{he2022synthetic}, fall short in precisely assessing dense annotations, while detector-based metrics~\cite{suri2023gen2det} struggle to capture detailed descriptions and finer relationships between objects. At the layout level, existing methods are constrained by predefined rules~\cite{NEURIPS2023_f8ad010c} and often neglect to consider other reasonable candidates~\cite{inoue2023layoutdm}.
As illustrated in Fig.~\ref{fig:method-comp}, current methods face limitations due to costly ground truth annotations and a lack of effective quality assessment metrics.
To solve these issues, several essential questions are raised: 1) How can we reduce the reliance on annotations, including caption and layout, while aligning closely with real data distributions? 
2) How do we measure the quality for multiple instances, and can we propose a new metric to select good ones? 
3) Does the proposed metric reflect the final downstream performance when used for training? 

% --------------------- 
% ---------------------

To this end, we propose Auto Cherry-Picker (ACP), a data generation pipeline leveraging pre-trained generative models and LLMs to generate images, detailed descriptions, and layout annotations for cross-modality perception and reasoning tasks. 
ACP comprises a raw data generator and a comprehensive data filter. 
The raw data generator redefines the current data synthesis paradigm by minimizing dependence on predefined annotation from original training data and instead using object combinations in natural language.
Leveraging these combinations, we first use LLMs to sample fine-grained scene graphs, including object attributes, relations, captions, and spatial layouts.
Next, controllable T2I models generate images based on these scene graphs.
This pipeline enables scalable synthetic data generation while aligning with real data distributions through object combinations rooted in real-world data priors.
It also addresses unbalanced distribution issues by adjusting the category proportion, especially in long-tailed scenarios. 
To ensure the quality of synthetic data, we introduce a comprehensive metric, \textbf{C}omposite \textbf{L}ayout and \textbf{I}mage \textbf{S}core (CLIS), in data filter.
CLIS evaluates the reasonableness of generated layouts (CLIS-L) and the quality and alignment of generated images (CLIS-I).
CLIS-L assesses the similarity between generated and ground truth layouts from data priors. 
Fig.~\ref{fig:teaser}(a,b) shows that high-quality layouts assessed by CLIS-L mirror real-world layouts and are more likely to yield high-quality images. 
CLIS-I evaluates visual quality and alignment with scene graphs, using priors from pre-trained large-scale models.
Fig.~\ref{fig:teaser}(c, d) shows that images with high CLIS-I exhibit superior visual quality and strong alignment with their corresponding scene graph.
The filtered scene graphs and corresponding images are used as training samples.

Through comprehensive evaluations, CLIS significantly enhances the performance of the state-of-the-state generation model, InstanceDiffusion, across various generation perspectives, including image fidelity, alignment to text, and layout control. 
Additionally, we observe a positive correlation between CLIS and performance gains in downstream tasks.
Scaling up the training samples generated by ACP results in substantial performance gains in perception and reasoning tasks, particularly in long-tail and open-vocabulary scenarios. 
On the LVIS dataset~\cite{gupta2019lvis}, we observe a +6.0\% improvement in AP$_{r}^{mask}$ for the long-tail setting using Mask R-CNN~\cite{he2017mask} and a +1.3\% gain in AP$_{novel}^{box}$ for the open-vocabulary setting using Grounding DINO~\cite{liu2023grounding}. 
Additionally, we achieve a score of +80.1 on the MME benchmark and an accuracy improvement of +0.4 on the GQA benchmark with LLaVA-v1.5~\cite{liu2024visual}, validating its efficiency in multi-modal perception and reasoning settings.

We summarize our technical contributions as follows:
\begin{enumerate}[label={\bf {{$\bullet$}}}, leftmargin=*, topsep=0.5ex, itemsep=-0.5ex, partopsep=0.75ex, parsep=0.75ex, partopsep=0pt, wide, labelindent=0pt]
    \item We propose ACP, an innovative training data generator for cross-modality perception and reasoning tasks. It is scalable and aligns with real data distributions.
    \item We design a comprehensive metric, CLIS, to filter generated data effectively by assessing layouts and images based on priors from real data or pre-trained large-scale models.
    \item Extensive experiments on visual and cross-modality perception and reasoning benchmarks demonstrate that ACP enhances model performance across various downstream tasks. The correlation between CLIS and performance gain among downstream tasks is well studied.
\end{enumerate}
\section{Related Work}
\label{sec:related_work}

\noindent
\textbf{Text to Image Generation.} Diffusion-based 
approaches~\cite{nichol2021glide,ramesh2022hierarchical,rombach2022high,saharia2022photorealistic} generate images as iterative denoising steps from random noise.
Stable Diffusion~\cite{rombach2022high} performs diffusion steps in the latent space of pre-trained autoencoders to achieve efficient training and sampling.
Subsequent studies extend text-to-image diffusion models with layout controllability by introducing auxiliary input signals~\cite{li2023gligen, zhang2023adding} or spatial tokens~\cite{yang2023reco} during training.
Another line of works~\cite{bar2023multidiffusion,chen2023trainingfree,g2023zero,singh2023high,Xie_2023_ICCV} follows a training-free approach by directly intervening the cross-attention layers during the sampling process.
However, these controllable T2I methods rely on visual annotations during inference.
With recent progress in the field of LLMs, LLM has been introduced in the T2I system to enhance text understanding and alignment~\cite{feng2024layoutgpt,gani2023llm,lian2024llmgrounded,qu2023layoutllm,yang2024mastering,wu2024selfcorrecting}, further extending methods to condition on detailed text descriptions without the need for manually designing layouts.
LMD~\cite{lian2024llmgrounded} and LayoutGPT~\cite{feng2024layoutgpt} use LLMs as text-guided layout generators through in-context learning.
SLD~\cite{wu2024selfcorrecting} uses LLM to correct the misalignment between the generated images and the user prompt in an iterative closed-loop process.
RPG~\cite{yang2024mastering} further adopts the recaptioning and planning of MLLM for subregion generation. 
Compared to previous works, we extend the generation paradigm to be conditioned on simple object combinations in natural language, broadening its applicability in scaling up synthetic data.

\noindent
\textbf{Learning from Synthetic Data.}
Deep learning models, especially for dense prediction tasks, typically require large amounts of data, which can be costly.
Therefore, many works use synthetic data to approximate information gathered or measured in the real world~\cite{wang2019learning,wood2021fake}. 
Synthesizing training samples with dense annotations is conditioned on various references.
Some works~\cite{chen2023geodiffusion,liu2019pixel,tan2023diffss} utilize the layout-to-image paradigm to synthesize training samples, conditioning on visual annotations like segmentation masks or bounding boxes. 
Some~\cite{li2023open,wu2023datasetdm} utilize an off-the-shelf perception model or adopt a perception head to obtain dense annotations of synthetic images generated based on detailed text descriptions.
A series of works~\cite{ge2022dall,zhao2022x} can condition on objects via copy-paste synthesis pipeline. However, the reasonableness of layouts is not considered.
Among all these studies, no works explore a fully reasonable language-driven pipeline.
To fill this gap, our method is driven purely by language and does not require expensive, manually annotated dense labels. 

%--------------------------------- Method Image ---------------------------------%
\begin{figure*}[t]
    \centering
    \includegraphics[width=0.8\linewidth]{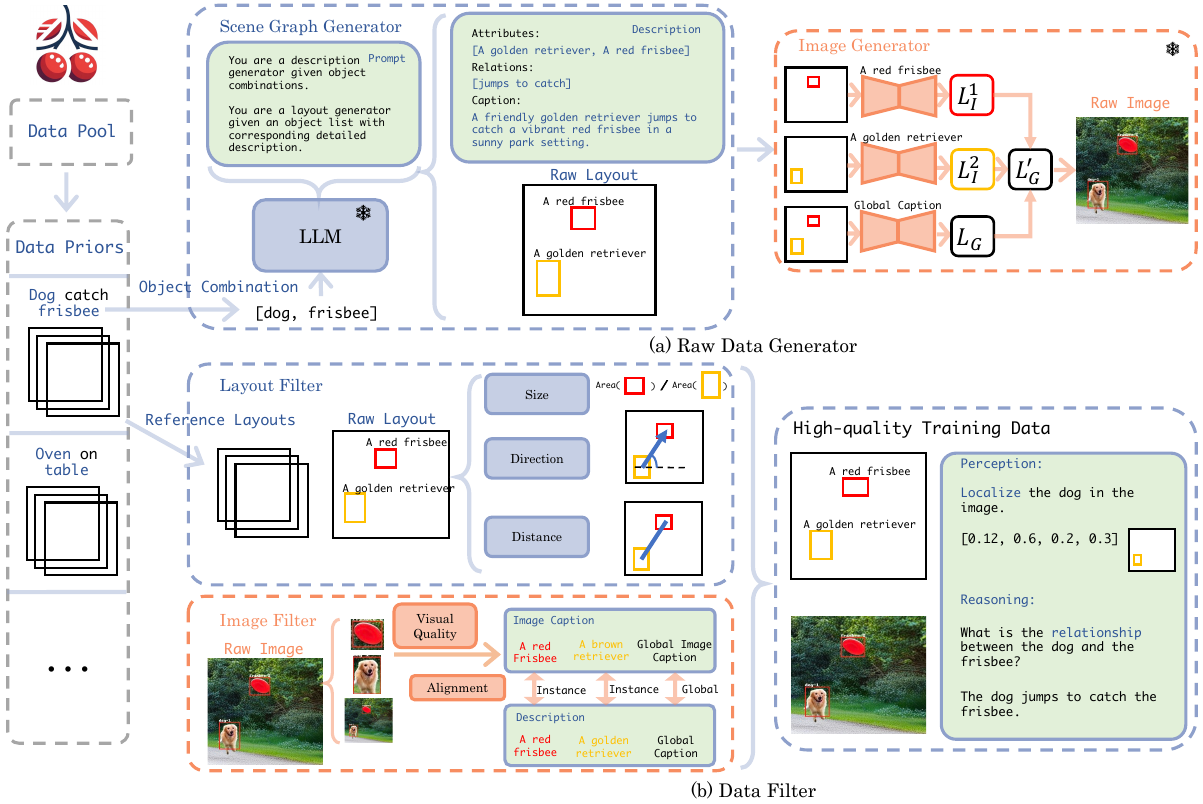}
    \caption{\footnotesize Illustration of Auto Cherry-Picker pipeline. It contains a (a) raw data generator and a (b) data filter using CLIS. Conditioned on input object combination sampled from data priors, Scene Graph Generator generates detailed attributes, relations, captions, and corresponding layouts. Subsequently, the Image Generator produces images based on the scene graph. These raw layouts and images are refined through filters using CLIS-L and CLIS-I, respectively, to produce high-quality training data.}
    \label{fig:arch}
    \vspace{-12pt}
\end{figure*}
%--------------------------------- Method Image ---------------------------------%

\noindent
\textbf{Generative Model Evaluation.}
Assessment of AI-generated content is challenging due to its subjective nature and the complexity of factors contributing to the generation quality.
Metrics like Inception Score (IS)~\cite{salimans2016improved}, Fréchet Inception Distance (FID)~\cite{heusel2017gans}, and LPIPS~\cite{zhang2018unreasonable} are commonly used for visual quality and diversity assessment. 
Some methods~\cite{hessel2021clipscore,chefer2023attend} focus on the alignment between text and generated image.
CLIPScore~\cite{hessel2021clipscore} computes the cosine similarity between text features and generated-image features extracted by CLIP. BLIP-CLIP~\cite{chefer2023attend} applies BLIP~\cite{li2022blip} to generate captions, then calculates the CLIP text-text cosine similarity between the generated captions and text prompts.
For layout quality assessment, LayoutDM~\cite{inoue2023layoutdm} proposes Maximum IoU~\cite{kikuchi2021constrained} to measure the similarity between generated and real layouts. They compute an optimal match between generated and real layouts, maximizing the average IoU for corresponding categories. 
HRS~\cite{bakr2023hrs} and T2I-CompBench~\cite{NEURIPS2023_f8ad010c} evaluate fixed spatial relationships, such as right, bottom, near, etc., of layouts using simple rules.
Compared to these metrics, CLIS evaluates instance-level results in complex scenes and combines the reasonableness of layout and visual quality in one shot, making it a suitable metric to generate high-quality data for downstream tasks.

\section{Method}
\label{sec:method}

Auto Cherry-Picker is a training-free cross-modality perception and reasoning dataset generation pipeline.
It can produce pairs of images and scene graphs conditioned on object combinations while automatically selecting high-quality ones for training downstream models. 
We first introduce the preliminary in Sec.~\ref{sec:preliminary}. Then, we detail our framework, including the raw data generator and the data filter in Sec.~\ref{sec:fram}. Finally, we explain the deployment on various downstream tasks in Sec.~\ref{sec:depl}.

\subsection{Preliminary}
\label{sec:preliminary}
\noindent
\textbf{Task Formation.}
Given a data pool $D_P=\{D_t\}_{t=1}^{T}$, where $D_t$ denotes the training set for downstream task $t$, the objective is to generate a high-quality synthetic dataset $\mathcal{D}_t$ that, when combined with original $D_t$, enhances model performance on the corresponding downstream task $t$.

\noindent
\textbf{Data Priors.}
Data priors are defined as
\begin{equation}
\label{eq:data-priors}
\small
    P = \{p_i=((s_i,o_i),r_i,L_i)|L_i=\{(l_s,l_o)_k^i\}_{k=1}^{M_i}\}_{i=1}^{N}
\end{equation}
where $s_i$ denotes the subject list, $o_i$ denotes the object list, $(s_i,o_i)$ denotes the object combination, $r_i$ represents the relationship between them, $L_i$ is a set of ground truth layouts corresponding to $(s_i,o_i)$ and $r_i$. 
We use LLMs to extract layouts, corresponding categories, and relationships from open-source datasets $D_P$, resulting in data priors $P$.

\noindent
\textbf{Synthetic Data.}
A single item $d_i$ in synthetic dataset $\mathcal{D}$ is defined as
\begin{equation}
\small
    d_i = \{(G_i,I_i) \mid G_i=(c_i,\{(\mathcal{S}_k,\mathcal{O}_k)\}_{k=1}^{K_i})\}
\end{equation}
\begin{equation}
\small
   (\mathcal{S}_k,\mathcal{O}_k) = \{(s_k, o_k), (a_s, a_o)_k, (l_s,l_o)_k, r_k\}
\end{equation}
where $I_i$ denotes the image, and $G_i$ denotes the scene graph. $G_i$ includes an overall caption $c_i$ and a set of detailed instance-level annotations $(\mathcal{S}_k,\mathcal{O}_k)$. Each pair includes original labels $(s_k,o_k)$, attributes $(a_s, a_o)_k$, layouts $(l_s, l_o)_k$, and their relationship $r_k$.

\subsection{ACP Framework}
\label{sec:fram}
As depicted in Fig.~\ref{fig:arch}, we aim to design a high-quality cross-modality training data generator comprising two key components:  a raw data generator and a data filter. The former seeks to generate data, while the latter selects good ones.

\noindent
\textbf{Raw Data Generator.} 
ACP generates data samples by harnessing information from data priors $P$ to 1) closely align with real data distributions and 2) enable scalable generation.
As shown in Fig.~\ref{fig:arch}(a), to generate $d_i$, the generator first samples object combinations $\{(s_k,o_k)\}_{k=1}^{K_i}$ from $P$. Next, LLMs generate descriptions from these initial combinations, leveraging their in-context learning capability~\cite{brown2020language}. 
The description contains detailed attributes $\{(a_s,a_o)_k\}_{k=1}^{K_i}$, relationships $\{r_k\}_{k=1}^{K_i}$ between different objects, and an overall dense caption $c_i$. 
Then, we utilize the spatial reasoning ability of LLMs to plan layouts based on the relationships and descriptions of objects. 
The LLM involved in this process is referred to as the scene graph generator.
Using the synthetic detailed description and spatial layouts, we adopt an off-the-shelf diffusion-based image generator to generate images by initiating the reverse diffusion process with different random noise. 

% --------------------
\begin{figure*}[t]
    \centering
    \includegraphics[width=0.95\linewidth]{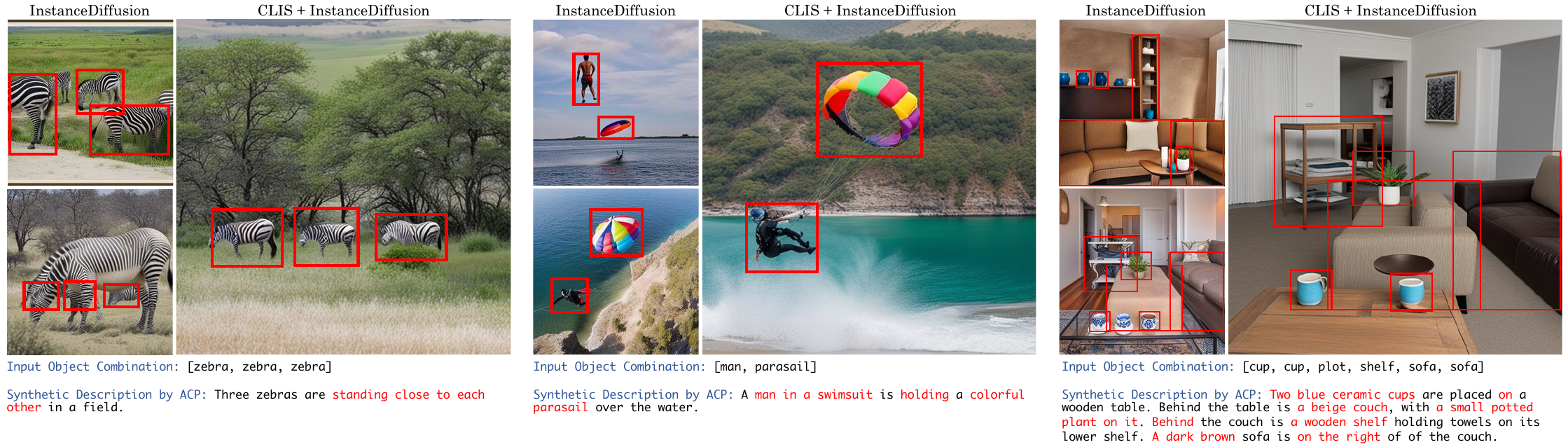}
    \vspace{-8pt}
    \caption{\footnotesize Comparison of generation results based on the same input object combinations and synthetic descriptions with and without CLIS. More generation results can be found in Appx.~\ref{app:vis}.
    % Visualization examples of CLIS based on InstanceDiffusion.
    }
    \label{fig:vis-comp}
    \vspace{-12pt}
\end{figure*}
% --------------------

\begin{table*}[t]
\begin{minipage}[t!]{0.45\textwidth}
\centering
\scriptsize
\begin{tabular}{l|ccc}
    \toprule[0.1em]
    Model & FID$\downarrow$ & CLIP score$\uparrow$ & YOLO score$\uparrow$\\
    \midrule
    Stable Diffusion~\cite{rombach2022high} & 56.8 & 26.6 & N/A\\
    \midrule
    BoxDiff-SD~\cite{Xie_2023_ICCV} & 60.0 & 26.4 & 4.4 \\
    BoxDiff-GLIGEN~\cite{Xie_2023_ICCV} & 61.0 & 25.9 & 21.5\\
    GLIGEN~\cite{li2023gligen} & 63.5 & 25.4 & 35.1 \\
    InstanceDiffusion~\cite{wang2024instancediffusion} & 53.5 & 25.2 & 45.6 \\
    \midrule
    GLIGEN w. CLIS & 59.9 (-3.6) & 25.8 (+0.4) & 36.8 (+1.7)\\
    InstanceDiffusion w. CLIS & 48.9 (-4.6) & 25.8 (+0.6) & 47.9 (+2.3)\\
    \bottomrule[0.1em]
\end{tabular}
\captionof{table}{\footnotesize Generation results of CLIS on the COCO val set.}
\label{tab:gen}
\end{minipage}
\hfill
\begin{minipage}[t!]{0.54\textwidth}
\centering
\includegraphics[width=\textwidth]{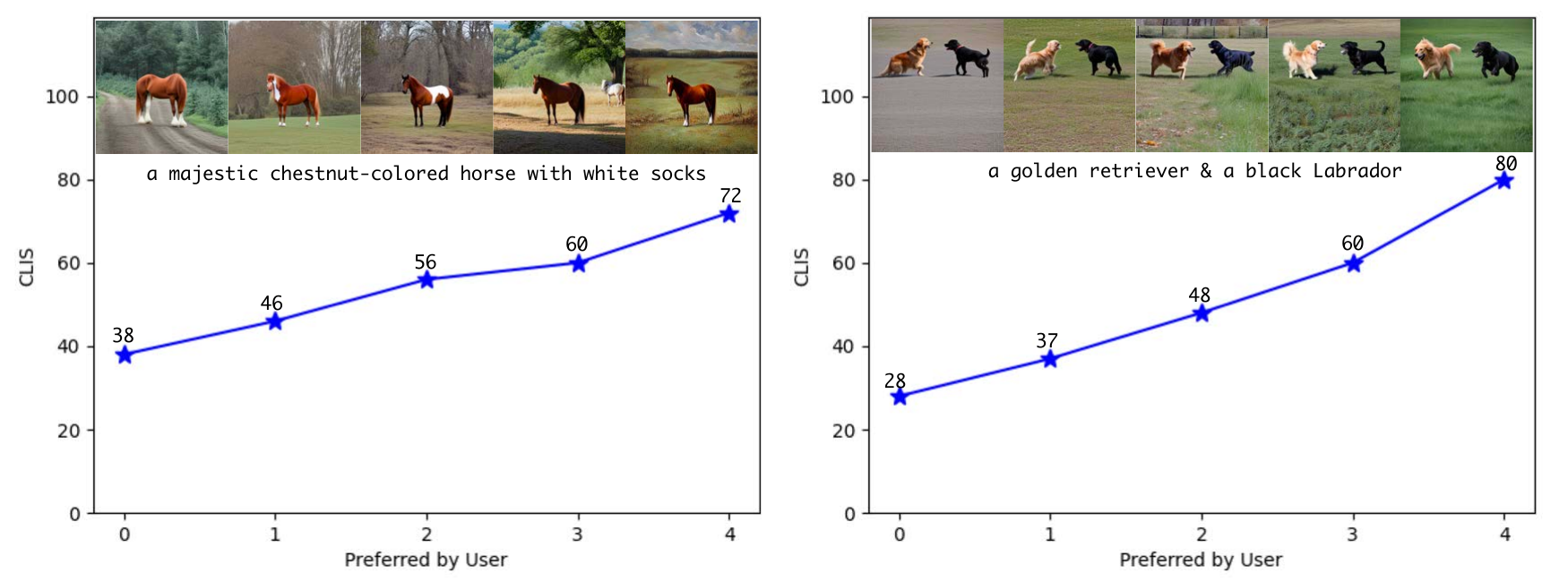}
\vspace{-18pt}
\captionof{figure}{\footnotesize Consistent with human judgement. See details in Appx.~\ref{appx:consistency}.}
\label{fig:con}
\end{minipage}
\vspace{-12pt}
\end{table*}

\noindent
\textbf{Data Filter.}
For the raw generated data from stage one, a data filter is utilized to cherry-pick high-quality training data. As depicted in Fig.~\ref{fig:arch}(b), the Layout Filter with CLIS-L and the Image Filter with CLIS-I assess the quality of layouts and image contents separately.

\noindent
\textbf{(a) Layout Filter.}
Precise relationships between objects in detailed image captions are essential for reducing hallucinations~\cite{zhai2024hallecontrolcontrollingobjecthallucination} and enhancing the reasoning abilities of MLLMs. 
Layout attributes provide important relational information, serving as valuable supplements for image-level evaluation.
High-quality layouts are also crucial for improving the likelihood of image generators generating high-quality images.
Since synthetic layouts lack ground truth, data priors are used to assess the reasonableness of layout pairs. 
Intuitively, we assume that a layout pair similar to the existing high-quality layout pairs is itself of high quality. 
CLIS-L evaluates this similarity from three perspectives: size, distance, and direction. 
Formally, given an object combination $(s, o)$, their relationship $r$, and corresponding layout pair $\mathcal{L}_1=(l_s, l_o)$, the reference layouts from data priors can be denoted as $L_P=P((s,o),r)$, CLIS-L is defined as:
\begin{equation}
\small
\begin{aligned}
\label{equ:clis-l}
    \text{CLIS-L}(s,o,r,\mathcal{L}_1)=F_{p}(\{S_{size}(\mathcal{L}_1,\mathcal{L}_2)+\\S_{dist}(\mathcal{L}_1,\mathcal{L}_2)+S_{dir}(\mathcal{L}_1,\mathcal{L}_2)\}|{\mathcal{L}_2 \in L_P})
\end{aligned}
\end{equation}
Here, $F_{p}$ refers to the operation of obtaining the $p$-th percentile to make CLIS-L robust against potential errors in $P$. $S_{size}$ and $S_{dist}$ are computed using the following similarity function $S_{sim}$:
\begin{equation}
\small
    S_{sim}(\mathcal{L}_1,\mathcal{L}_2,t) = 1 - \frac{\left|f_{t}(\mathcal{L}_1) - f_{t}(\mathcal{L}_2)\right|}{max(f_{t}(\mathcal{L}_1), f_{t}(\mathcal{L}_2))} 
\end{equation}
where $t\in\{\text{Area},\text{IoU},\text{RD}\}$ denotes the attribute $f_t$ focuses on. Specifically, $f_{\text{Area}}$ computes the area ratio of layout pair, while $f_{\text{IoU}}$ and $f_{\text{RD}}$ compute the IoU and relative distance, respectively. 
Thus, we define $S_{size}$ and $S_{dist}$:
\begin{equation}
    \label{equ:size}
\small
    S_{size}(\mathcal{L}_1,\mathcal{L}_2) = S_{sim}(\mathcal{L}_1,\mathcal{L}_2,\text{Area})
\end{equation}
\begin{equation}
\label{equ:dis}
\small
    S_{dist}(\mathcal{L}_1,\mathcal{L}_2) = S_{sim}(\mathcal{L}_1,\mathcal{L}_2,\text{IoU}) + S_{sim}(\mathcal{L}_1,\mathcal{L}_2,\text{RD})
\end{equation}
For $S_{dir}$, we compute the cosine similarity between direction vectors of $\mathcal{L}_1$ and $\mathcal{L}_2$:
\begin{equation}
    \label{equ:dir}
\small
    S_{dir}(\mathcal{L}_1,\mathcal{L}_2) = cos(f_{\text{Dir}}(\mathcal{L}_1), f_{\text{Dir}}(\mathcal{L}_2))
\end{equation}
where $f_{\text{Dir}}$ computes the direction vector between two layouts.

\noindent
\textbf{(b) Image Filter.}
To enable models with strong perception and complex reasoning abilities, we emphasize the visual quality of the image and its alignment with corresponding detailed descriptions. 
A pre-trained image captioning model generates global and local descriptions for synthetic images, respectively, where clear descriptions indicate high image quality and fidelity.
Local descriptions are generated by guiding the model to focus on the corresponding layouts.
An LLM then compares these descriptions with those in the generated scene graph to evaluate alignment.
Thus, CLIS-I is formulated as:
\begin{equation}
\label{equ:image-metric}
\small
    \text{CLIS-I}(G,I) = F_{sim}(G'^T, G^T) = F_{sim}(F_C(I, G^L), G^T)
\end{equation}
where $G^T$ represents the textual part of the scene graph $G$ (global caption and local objects attributes), $G^L$ represents the layouts within $G$, $I$ is the synthetic image corresponding to $G$, and $F_C$, $F_{sim}$ denote the captioning and similarity functions, respectively. 
Notably, CLIS-I includes instance-level scores without additional calculations, enabling instance-level filtering.
Please refer to Appx.~\ref{app:fundation-models-and-prompt} for details of foundation models and prompts utilized in ACP.

\subsection{Deployment on Downstream Tasks}
\label{sec:depl}
Cross-modality data generated by ACP can be readily transformed into training samples for various downstream tasks. In this work, we validate its generalizability in two primary settings: visual perception and multi-modal perception and reasoning.

\noindent
\textbf{Visual Perception Tasks.}
Layouts naturally serve as bounding boxes for detection tasks in synthetic training samples.
To further deploy these samples for segmentation tasks, we adopt SAM~\cite{kirillov2023segment} to obtain masks within the layout.
Combining these segmentation masks with layout annotations and images prepares synthetic samples for visual perception tasks, especially in unbalanced scenarios such as long-tailed instance segmentation and open-vocabulary object detection.

\noindent
\textbf{Multi-modal Perception and Reasoning Tasks.}
For instruction fine-tuning of MLLMs, we construct question-answer pairs using a predefined question template. Details can be found in Appx.~\ref{app:templates-for-multi-modal-downstream-tasks}.
The synthetic instruction data aims to enhance the perception and reasoning abilities of MLLMs. 
The question template includes prompts that require models to provide detailed descriptions of objects based on location or category, localize objects from descriptions, distinguish relationships between objects, and more.

\noindent
\textbf{CLIS Setting.} 
CLIS selects high-quality samples through two approaches: 1) it chooses the highest-scoring image and layouts from a set generated from the same object combinations, and 2) further to refine high-quality training samples for different downstream tasks, it applies independent score thresholds for layouts (CLIS-L), instances, and images (CLIS-I). Please see Appx.~\ref{app:clis-settings} for more details about CLIS computation, score distribution, and filtering ratio.

% --------------------
\begin{figure}[t]
    \centering
    \includegraphics[width=.95\linewidth]{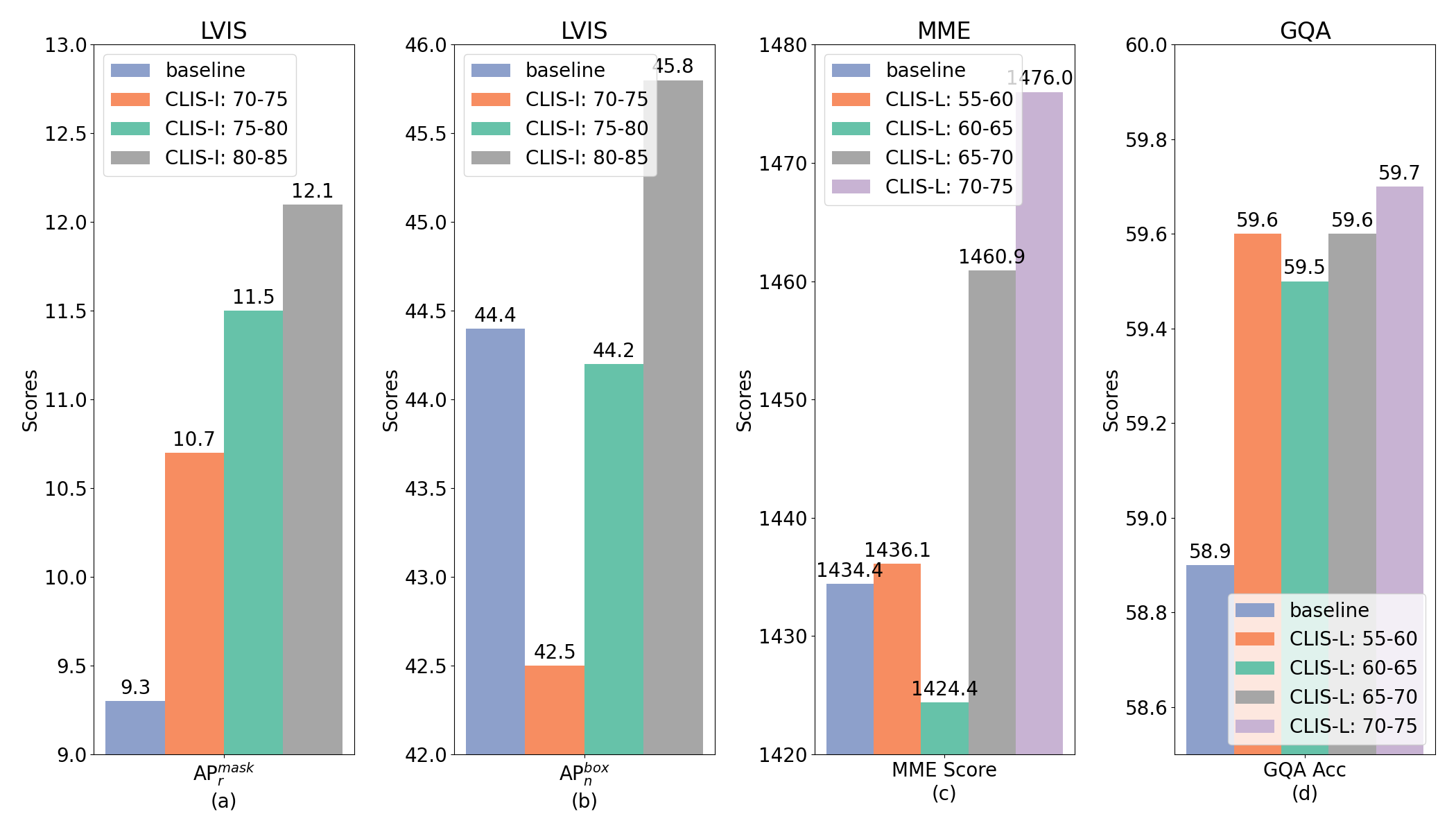}
    \vspace{-8pt}
    \caption{\footnotesize Correlation between CLIS and performance gains on downstream tasks. (a,b): Synthetic data with different ranges of CLIS-I on long-tailed instance segmentation and open-vocabulary detection scenarios of LVIS benchmarks using Mask R-CNN and Grounding-DINO as baseline, respectively. (c,d): Synthetic data with different ranges of CLIS-L on multi-modal perception and reasoning MME and GQA benchmarks based on LLaVA-v1.5.
    }
    \label{fig:cor}
    \vspace{-12pt}
\end{figure}
% --------------------

\section{Experiments}
\label{sec:exp}

\begin{table*}[!t]
    \footnotesize
        \centering
    \caption{Results on visual perception downstream tasks. (left): LVIS long-tailed instance segmentation benchmarks. (right): Open-vocabulary object detection benchmarks.}
    \label{tab:long-tail}
    \label{tab:open-vocab}
    \resizebox{0.52\linewidth}{!}{
        \begin{tabular}{cc|cc}
        \toprule[0.1em]
        Method & Backbone & AP$_{r}^{mask}$ & AP$^{mask}$ \\
        \midrule
        Mask R-CNN~\cite{he2017mask} & ResNet-50 & 9.3 & 21.7 \\
        w. ACP & ResNet-50 & 14.5 (+5.2) & 22.8 (+1.1) \\
        \midrule
        CenterNet2 w. Copy-Paste~\cite{ghiasi2021simple} & Swin-B & 29.3 & 39.3 \\
        w. ACP & Swin-B & 30.7 (+1.4) & 39.6 (+0.3) \\
        \bottomrule[0.1em]
        \end{tabular}
    }\hfill
    \resizebox{0.46\linewidth}{!}{
        \begin{tabular}{cccc|cc}
        \toprule[0.1em]
        Dataset & Method & Backbone & AP$_{novel}^{box}$ & AP$^{box}$ \\ 
        \midrule
        % Grounding-DINO & Swin-T & zero-shot & 23.5 & 31.9 \\ 
        \multirow{2}{*}{LVIS} & Grounding-DINO & Swin-T  & 31.7 & 48.7 \\
        & w.ACP & Swin-T  & 33.0 (+1.3) & 49.2 \\ % (+0.5)  
        \midrule
        \multirow{2}{*}{COCO} & Grounding-DINO & Swin-T  & 60.4 & 57.1 \\
        & w.ACP & Swin-T  & 60.8 (+0.4) & 56.9 \\ 
        \bottomrule[0.1em]
        \end{tabular}
    }
    \vspace{-12pt}
\end{table*}

\begin{table}[!t]
    \footnotesize
    \centering
    \caption{\footnotesize ACP boosting the results on the multi-modal MME and GQA benchmarks.}
    \label{tab:multi}
    \resizebox{0.9\linewidth}{!}{
        \begin{tabular}{cc|cc}
        \toprule[0.1em]
        Method & LM Backbone & MME & GQA \\
        \midrule
        LLaVA-1.5 & Vicuna-7B & 1434.4 & 58.9 \\
        LLaVA-1.5 & Vicuna-13B & 1438.3 & 60.7 \\
        LLaVA-1.5 & LLama-3-8B & 1445.3 & 60.1 \\
        \midrule
        LLaVA-1.5 w. ACP & Vicuna-7B & 1514.5 (+80.1) & 59.3 (+0.4) \\
        \bottomrule[0.1em]
        \end{tabular}
    }
    \vspace{-12pt}
\end{table}

\noindent
We first introduce experiments set up in Sec.~\ref{sec:imp}. We then validate CLIS in Sec.~\ref{sec:stu-eff} from two perspectives: 1) its correlation with image fidelity of generated samples and 2) its correlation with the performance gain in downstream tasks.
Next, we demonstrate ACP's effectiveness on several downstream tasks and demonstrate its potential for continuous scaling up of data size in Sec.~\ref{sec:syn}.
We validate the designed modules in ACP via a series of ablation studies in Sec.~\ref{sec:abla}.
More results are available in the supplementary material. All the source code and datasets will be available to the public.

\subsection{Implementation Details}
\label{sec:imp}
\noindent
\textbf{Datasets.} 
We evaluate generation quality using COCO~\cite{lin2014microsoft} following prior works~\cite{g2023zero,dong2023dreamllm}. 
We randomly sample 1181 images from the COCO validation set, each paired with a fixed caption. 
For downstream tasks, we conduct object detection and instance segmentation experiments on COCO and LVIS v1.0~\cite{gupta2019lvis}. 
Additionally, we evaluate image-based visual question answering (VQA) using the MME~\cite{fu2024mme} and GQA~\cite{hudson2019gqa} benchmarks. The MME Perception benchmark is widely used to evaluate the perception abilities of MLLMs. GQA is a comprehensive dataset for assessing visual reasoning abilities.

\noindent
\textbf{Baselines.}
For generation models, we primarily select existing controllable diffusion-based T2I models, including GLIGEN~\cite{li2023gligen}, BoxDiff~\cite{Xie_2023_ICCV} and InstanceDiffusion~\cite{wang2024instancediffusion}, following their official settings. We also include Stable Diffusion~\cite{rombach2022high} as a baseline T2I model. 
For downstream tasks, we adopt Mask R-CNN~\cite{he2017mask} and CenterNet2~\cite{zhou2021probabilistic} as baselines for long-tailed instance segmentation. For open-vocabulary object detection, we use Grounding-DINO~\cite{liu2023grounding,zhao2024open}, and for VQA, we employ LLaVA-v1.5~\cite{liu2023improved,liu2024visual}.

\noindent
\textbf{Evaluation Protocols.} 
To assess image-level quality, we employ the Fréchet Inception Distance (FID)~\cite{heusel2017gans} with the COCO validation set as the reference dataset.
We also use the CLIP score and YOLO score~\cite{li2021image} to measure alignment and layout accuracy, respectively.
For segmentation and object detection, we use Average Precision (AP) as the primary evaluation metric
and report AP for novel and rare categories in open-vocabulary and long-tailed scenarios, respectively. For VQA benchmarks, we report the averaged score for MME and accuracy for GQA.
Please refer to Appx.~\ref{app:exp-setup} for more details about the setup of our experiments.

% --------------------
\begin{figure}[t]
    \centering
    \includegraphics[width=\linewidth]{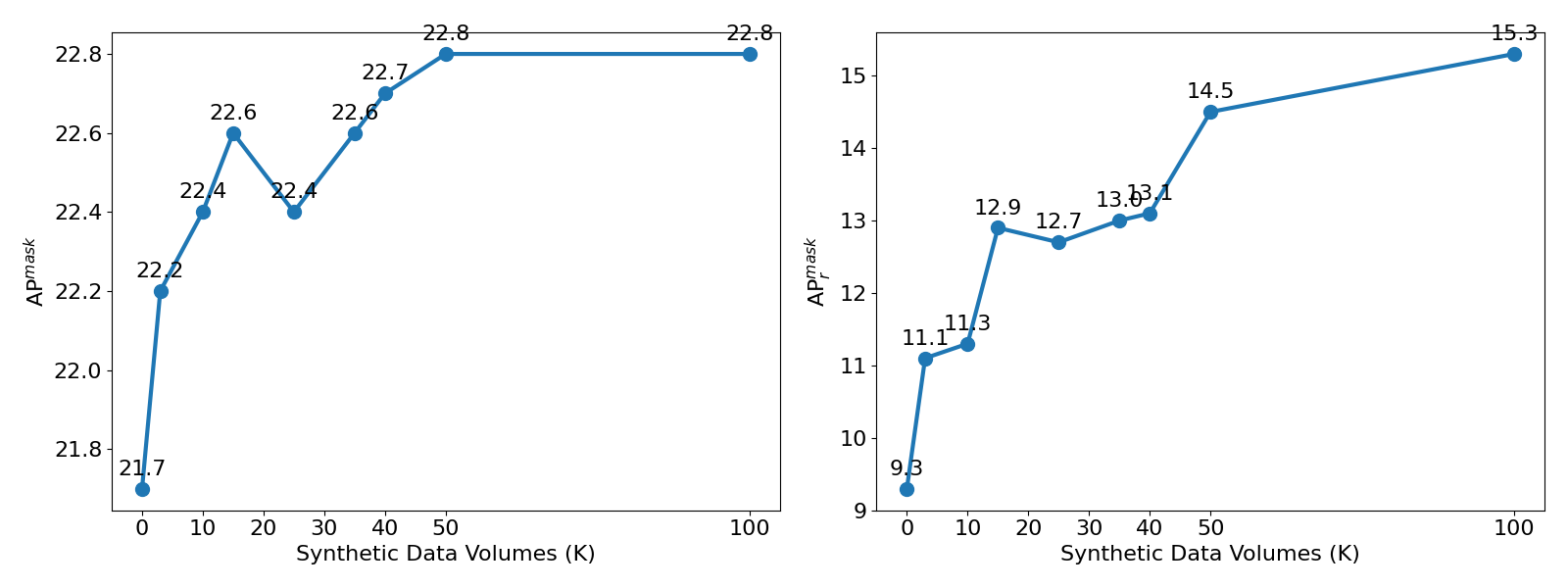}
    \captionof{figure}{\footnotesize Data scaling on the LVIS benchmark using Mask R-CNN.}
    \label{fig:scale}
    \vspace{-12pt}
\end{figure}
% --------------------

\subsection{Study Efficacy of CLIS}
\label{sec:stu-eff}
\noindent
\textbf{Generation Results.} 
We first evaluate the efficacy of CLIS from a conventional generative perspective.
Table~\ref{tab:gen} shows that CLIS enhances generation quality, as evidenced by a decrease in FID scores and increases in both CLIP and YOLO scores when applied to both GLIGEN~\cite{li2023gligen} and InstanceDiffusion~\cite{wang2024instancediffusion}, demonstrating its effectiveness and generalizability. 
We also present visualization results in Fig.~\ref{fig:vis-comp}, comparing samples generated with and without CLIS, using the same input object combinations and synthetic descriptions. CLIS enhances layout quality in both action-based and complex multi-object scenarios. Furthermore, it ensures better visual quality and alignment between images and textual descriptions.  
We additionally evaluate its consistency with human preference in Appx.~\ref{appx:consistency}.

\noindent
\textbf{Correlation with Performance Gains on Downstream Tasks.} 
We first compare CLIS with other widely used selection metrics. ACP generates 30K raw training samples for long-tailed instance segmentation. We then apply three different selection methods (random, CLIP, and CLIS) to pick the top 50\% of synthetic data to train Mask R-CNN. Fig.~\ref{fig:teaser}(e) shows that CLIS achieves the highest score on LVIS, demonstrating its superior correlation compared to other metrics.

We further analyze the separate correlation of CLIS-I on visual perception task performance and CLIS-L on multi-modal perception and reasoning tasks.
We sample an equal number of synthetic samples across different CLIS-I and CLIS-L ranges to train baseline models, using 10K samples for CLIS-I and 1.3K for CLIS-L. 
Fig.~\ref{fig:cor} shows that higher CLIS-I correlates with improved performance in both long-tailed and open-vocabulary settings. Similarly, a positive correlation is observed between CLIS-L and performance on the MME and GQA benchmarks.
These results validate the effectiveness of using CLIS in data filter across various downstream tasks.

\subsection{Synthetic Dataset Scale Up}
\label{sec:syn}
\noindent
\textbf{Data Scaling.}
To investigate the data scaling effects, we conduct experiments on LVIS using Mask R-CNN as the baseline, varying the size of synthetic data. Fig.~\ref{fig:scale} illustrates that ACP consistently boosts performance with increasing data volumes. 
Remarkably, ACP achieves its best results with the largest data size (100K), reaching an AP$_{r}^{mask}$ of 15.3 (+6.0) and an AP$^{mask}$ of 22.8 (+1.1), which shows potential for continuous scaling up.
Interestingly, we find that as the volume of selected data grows, performance initially rises rapidly while plateaus after roughly 50K synthetic examples. 
To balance computation efficiency with performance gains, we choose 50K as our default synthetic data volume for subsequent experiments.

\noindent
\textbf{Long-tailed Instance Segmentation Benchmark.}
Table~\ref{tab:long-tail} (left) presents our results on the long-tailed instance segmentation settings of LIVS.
ACP demonstrates significant performance gains over the commonly-used Mask R-CNN baseline, with an improvement of 1.1\% in $AP^{mask}$, and the most notable improvement in rare categories (+5.2\% AP$_{r}^{mask}$).
We also observe consistent performance improvements with a stronger CenterNet2 baseline, which employs Swin-B as the backbone and copy-paste~\cite{ghiasi2021simple} for data augmentation. With this setup, ACP achieves a 1.4\% higher AP$_{r}^{mask}$.
This underscores ACP's strong generalization ability across different detector architectures and its effectiveness in conjunction with existing data augmentation methods.

\noindent
\textbf{Open-vocabulary Object Detection Benchmark.}
We further demonstrate the effectiveness of ACP in the challenging open-vocabulary detection setting. 
We use Grounding-DINO~\cite{liu2023grounding} as our baseline, which is pre-trained on large-scale data corpus, including Objects365~\cite{shao2019objects365}, GoldG~\cite{kamath2021mdetr}, GRIT~\cite{peng2023kosmos}, and V3Det~\cite{wang2023v3det}, totaling 61.8M images, following~\cite{zhao2024open}. 
Table~\ref{tab:open-vocab} (right) shows that ACP performs favorably against Grounding-DINO by 1.3\% in LVIS AP$_{n}^{box}$ and 0.4\% in COCO AP$_{n}^{box}$, 
despite using a limited volume of generated training samples compared to the size of pre-train real data.
This validates how high-quality synthetic data can complement real data effectively.

\begin{table}[!t]
    \footnotesize
    \centering
    \caption{\footnotesize Compared with existing data generation methods on the LVIS benchmark. MosaicFusion defaults to using 4K synthetic images of rare categories. For a fair comparison, we extend it to all categories, denoted as MosaicFusion\textsuperscript{†}. Additionally, we adjust the rare ratio in ACP by incorporating 2K rare synthetic images while keeping the total number unchanged, denoted as ACP\textsuperscript{\ddag}.}
    % \vspace{-6pt}
    \label{tab:xp}
    \resizebox{0.98\linewidth}{!}{
        \begin{tabular}{c|ccc|ccc}
        \toprule[0.1em]
        Method & Backbone & AP$_r^{mask}$ & AP$^{mask}$ & Backbone & AP$_r^{mask}$ & AP$^{mask}$ \\
        \midrule
        CenterNet2 (baseline) & ResNet-50 & 17.8 & 26.1 & SwinB & 27.3 & 34.1 \\
        w. X-Paste~\cite{zhao2022x} & ResNet-50 & 17.9 & \underline{28.0} & SwinB & 26.8 & \underline{35.1}  \\
        w. MosaicFusion~\cite{xie2023mosaicfusion} & ResNet-50 & \underline{19.6} & 26.7 & SwinB & \underline{29.8} & 34.3 \\
        w. MosaicFusion\textsuperscript{†} & ResNet-50 & 18.6 & 27.0 & SwinB & 29.2 & 34.6 \\
        \midrule
        w.ACP & ResNet-50 & 19.2 & \underline{28.0} & SwinB & 29.6 & \textbf{35.2} \\ 
        w.ACP\textsuperscript{\ddag}& ResNet-50 & \textbf{21.8} & \textbf{28.1} & SwinB & \textbf{30.6} & \underline{35.1} \\
        \bottomrule[0.1em]
        \end{tabular}
    }
    \vspace{-12pt}
\end{table}

\noindent
\textbf{Multi-modal Benchmarks.}
We further evaluate the effectiveness of ACP on multi-modal perception and reasoning tasks.
We adopt LLaVA-v1.5 with Vicuna-7B as our baseline. 
Table~\ref{tab:multi} indicates that ACP significantly enhances the model's perception ability on the MME benchmark, achieving an improvement of 80.1, which exceeds the performance of LLaVA-v1.5 even with stronger language model backbones such as Vicuna-13B and LLama-3-8B.
Additionally, ACP improves performance on the widely recognized GQA reasoning benchmark. 
These results validate the effectiveness of our method in cross-modality settings.

\noindent
\textbf{Comparison with Previous Methods.}
We conduct a quantitative comparison with X-Paste~\cite{zhao2022x} and MosaicFusion~\cite{xie2023mosaicfusion} using CenterNet2 with two backbones, ResNet-50 and Swin-B.
Table~\ref{tab:xp} shows that ACP consistently achieves the highest AP$^{mask}$ on both backbones and strikes a strong balance between AP$^{mask}$ and AP$^{mask}_r$. 
When adjusting the rare ratio, ACP\textsuperscript{\ddag} can further enhance AP$^{mask}_r$ significantly.
In contrast, X-Paste yields high AP$^{mask}$ but provides no additional gains on AP$^{mask}_r$ compared to the baseline. Meanwhile, MosaicFusion boosts AP$^{mask}_r$ but offers only modest AP$^{mask}$ improvements, and its advantage in rare categories diminishes when extended to all categories. 
While ACP's AP$^{mask}$ improvements over X-Paste are relatively modest, ACP achieves significant gains in AP$^{mask}_r$, enhancing the robustness and practicality of trained detectors.
These results demonstrate the superiority of synthesizing images with reasonable layouts compared to composing training examples by pasting multiple synthesized instances onto a background.
Efficiency analysis can be found in Appx.~\ref{app:efficiency-effectiveness}.

\begin{table}[t]
    \footnotesize
    \centering
    \caption{\footnotesize Ablation study on Layout Generator. We compare our method to ACP with ground truth layouts on the LVIS using Mask R-CNN as our baseline. 
    }
    \label{tab:abla-layout}
    \resizebox{.75\linewidth}{!}{
    \begin{tabular}{l|c|cc}
        \toprule[0.1em]
        Model & Layout Generator & AP$_r^{mask}$ $\uparrow$ & AP$^{mask}$ $\uparrow$ \\
        \midrule
        Baseline & $\times$ & 9.3 & 21.7 \\
        \midrule
        ACP & $\times$ & 11.3 & 22.9 \\
        ACP & $\checkmark$ & 12.2 & 23.0 \\
        \bottomrule[0.1em]
    \end{tabular}
    }
    \vspace{-4pt}
\end{table}

\begin{table}[t]
    \footnotesize
    \centering
    \caption{\footnotesize Ablation study on CLIS-I. We compare three variants of CLIS-I by filtering images generated from the same scene graph.
    }
    \label{tab:abla-clis-i}
    \resizebox{.9\linewidth}{!}{
    \begin{tabular}{l|cc|c|cc}
        \toprule[0.1em]
        Model & $F_{C}$ & $F_{sim}$ & FID$\downarrow$ & AP$_r^{mask}$$\uparrow$ & AP$^{mask}$$\uparrow$ \\
        \midrule
        Baseline & $\times$ & $\times$ & N/A & 9.3 & 21.7 \\
        \midrule
        CLIP & $\times$ & $\checkmark$ & 40.8 & 10.3 & 21.9  \\
        CLIP-text & $\checkmark$ & $\times$ & 41.2 & 11.3 & 22.2  \\        
        CLIS-I & $\checkmark$ & $\checkmark$ & 41.3 & 11.5 & 22.3 \\
        \bottomrule[0.1em]
    \end{tabular}
    }
    \vspace{-12pt}
\end{table}

\subsection{Ablation and Analysis}
\label{sec:abla}
\begin{figure}[t]
    \centering
    \includegraphics[width=0.99\linewidth]{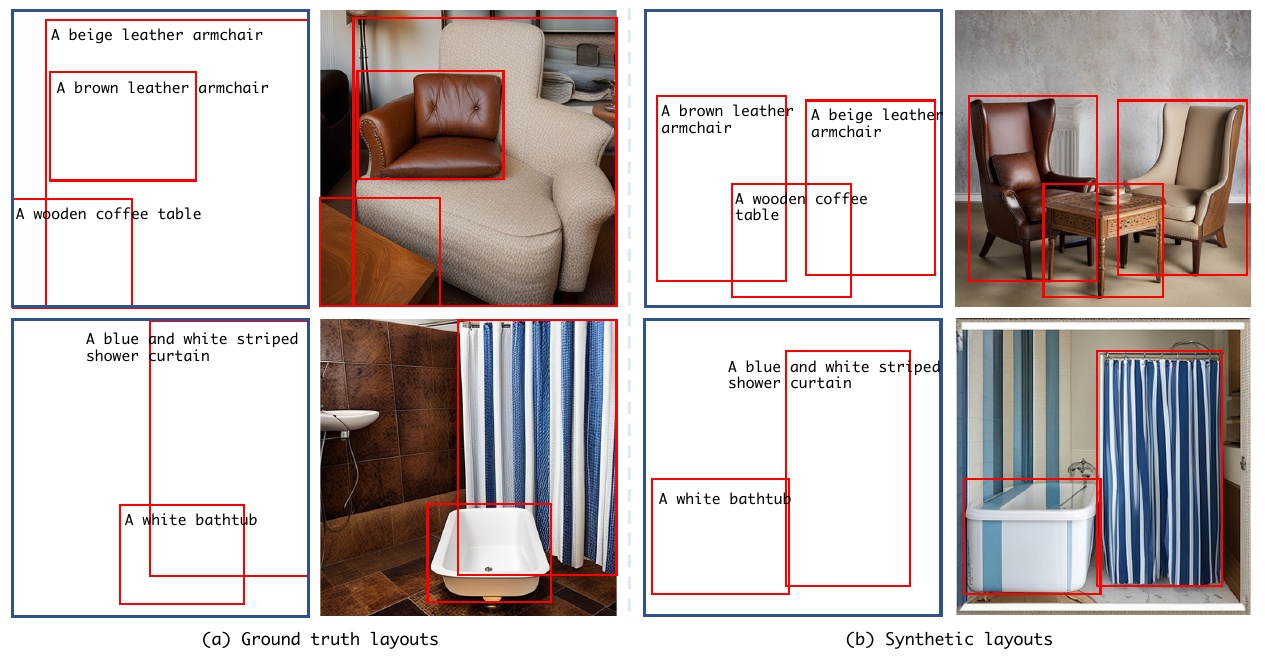}
    \caption{\footnotesize Visualization of the comparison between ground truth layouts and synthetic layouts, along with their corresponding synthetic images.}
    \label{fig:ablation-clis-l}
    \vspace{-12pt}
\end{figure}

\noindent
\textbf{Layout Generator.}
We perform an ablation study on the layout generator by comparing with ACP utilizing ground truth layouts derived from training annotations. 
First, we conduct a qualitative comparison by showcasing images generated from both ground truth layouts and synthetic layouts in Fig.~\ref{fig:ablation-clis-l}.
The layout generator is capable of producing high-quality, reasonable layouts and helps mitigate issues such as overlap and small object sizes in ground truth layouts, which can lead to sub-optimal results in diffusion models.
Next, we perform a quantitative comparison using 5K synthetic samples on LVIS with Mask R-CNN as our baseline. 
Table~\ref{tab:abla-layout} shows that ACP with synthetic layouts achieves a +0.9\% increase in AP$_r^{mask}$ compared to ACP with ground truth layouts, demonstrating the effectiveness of generated layouts.
We hypothesize that this improvement is due to the layout generator introducing data augmentation at the layout level rather than solely relying on the image generator's image-level augmentation while maintaining sufficient quality for the image generator.
Further examples can be found in Appx.~\ref{app:aba-lay}.

% --------------------
% --------------------

\noindent
\textbf{Components of CLIS-I.}
We conduct an ablation study on two operations in CLIS-I: $F_{sim}$ and $F_{C}$ in Eq~\ref{equ:image-metric}. 
Specifically, we replace LLM alignment function $F_{sim}$ with CLIP text alignment score and replace image caption function $F_{C}$ with CLIP score to directly measure alignment between image and text.
For the same set of scene graphs, the Image Generator produces 4 images for each scene graph, and these image filtering methods independently select the highest-scoring image.
Each method generates the same volume of data (4K) and uses the same scene graph annotations.
We evaluate these methods from both generation perspectives and performance gains in the downstream task. In particular, we use the LVIS benchmark and Mask R-CNN as the baseline detector.
Table~\ref{tab:abla-clis-i} shows that CLIS-I demonstrates the most significant performance gain in the downstream task, with a +2.2\% increase in $AP_r^{mask}$, aligning with our initial motivation.
Interestingly, while CLIP selection excels in generation evaluation, it yields suboptimal results in the downstream tasks.
This highlights that CLIS-I is more strongly correlated with downstream task performance gains than conventional generation metrics.

\section{Conclusion}
\label{sec:conclusion}
In this paper, we propose Auto Cherry-Picker, a cross-modality training data generator conditioned on object combinations with a comprehensively designed CLIS metric to ensure the quality of generated data. 
ACP is effective in various downstream tasks, including perception and reasoning tasks, particularly in improving the performance in annotation-scarce scenarios. 
CLIS can be used to pick high-quality generation samples, where we also find the generated data with higher CLIS can lead to better performance for perception tasks.
Moreover, our method can be easily adapted to stronger LLMs and image generation models. 
Our research bridges the gap between generation data and downstream performance.
We hope our results can inspire generation metric design in the future. 

\noindent
\textbf{Limitations.} Scaling the synthetic data size is resource-intensive. Currently, we only utilize a limited volume of synthetic data to ensure quality. Exploring methods that leverage low-quality data would be beneficial. We present more details on limitations and future work in the Appx.~\ref{app:limi}.
{
    \small
    \bibliographystyle{ieeenat_fullname}
    \bibliography{main}
}
\clearpage
\setcounter{page}{1}
\maketitlesupplementary

\section{Implementation Details of ACP}
\label{app:fundation-models-and-prompt}

In this section, we provide implementation details of ACP. Following our pipeline, we introduce details of Data Priors in Sec.~\ref{app:data-priors}, Scene Graph Generator in Sec.~\ref{app:scene-graph-generator}, Image Generator in Sec.~\ref{app:image-generator}, and Image Filter in Sec.~\ref{app:image-filter}.

\subsection{Data Priors}
\label{app:data-priors}

\noindent
\textbf{Data Priors Construction.} 
Data Priors $P$ is constructed from open-source datasets $D_P$, where each sample contains an image caption and associated annotations.
We use LLMs to parse each caption and extract object combination $(s_i,o_i)$ along with their relationship $r_i$, forming $\{(s_i,o_i),r_i\}_{i=1}^{N}$. 
For each extracted item $\{(s_i,o_i),r_i\}$, we retrieve the corresponding bounding boxes from annotations as $(l_s,l_o)^i_k$.
By iterating over the entire $D_P$, we gather all relevant layout pairs corresponding to $\{(s_i,o_i),r_i\}$ into a list $L_i$.
Thus, each data prior is formulated as $p_i=((s_i,o_i),r_i,L_i)$, aligning with Eq.~\ref{eq:data-priors}.

\noindent
\textbf{Reference Layouts Selection from Data Priors.}
Reference layouts selection is based on the object combination $(s, o)$ and their relationship $r$.
We extract reference layouts from $P$ that share the same category and relationship. 

\subsection{Scene Graph Generator}
\label{app:scene-graph-generator}

\noindent
\textbf{Choice of LLMs.}
We conduct experiments using a series of LLMs as the scene graph generator in our ACP pipeline on a limited data scale. Specifically, we employ Qwen-1.5-14B, Qwen-1.5-72B, and Qwen-1.5-110B to generate scene graphs. 
Each model produces 15K training examples from the same input object lists. 
These examples are then amalgamated with the original training data for COCO detection tasks. 
We apply them separately to a Mask R-CNN baseline under a standard 1$\times$ training schedule. 
Table~\ref{tab:app-dif} illustrates that the performance of different LLMs is comparable in downstream tasks. 
We opt for the smaller LLM, Qwen-1.5-14B, for the experiments described due to its faster inference speed.

\begin{table}[h]
    \begin{center}
    \caption{Different LLMs as scene graph generator in ACP on COCO detection task.}
    \label{tab:app-dif}
    \begin{tabular}{l|cc}
        \toprule[0.1em]
        Scene Graph Generator & AP$^{mask}$$\uparrow$ & AP$^{box}$$\uparrow$  \\
        \midrule
        Qwen1.5-14b & 34.5 & 37.8 \\
        Qwen1.5-72b & 34.5 & 37.8 \\
        Qwen1.5-110b & 34.2 & 37.7 \\
        \bottomrule[0.1em]
    \end{tabular}
    \end{center}
\end{table}

\noindent
\textbf{Prompts.}
\label{app:prom}
We provide our full prompts for the scene graph generator, including the description generator and layout generator.

\noindent
\textbf{Details of Layout Generation.}
We derive the input from the previous description and prompt LLMs to generate a layout for each object. The input follows a dictionary format, e.g., \{"objects": ["xx-1", "xx-2", "xx-3"], "caption": "xxx"\}, and the output is a list of dictionaries, e.g., [\{"object": "xx-1", "layout": [x,y,w,h]\}]. The Raw Layout in Fig~\ref{fig:arch}(a) represents the complete image layout, encompassing all object layouts for single-image generation.

\begin{center}
\begin{tcolorbox}
\textbf{Prompt for Description Generator}:\\
Task Description:\\
Your task is to generate a detailed description based on an object list.
The description should be a structured representation of a scene detailing its various elements and their relationships. The description consists of: 1. attributes of objects: The attributes should be descriptive of the color or texture of the corresponding object. 2. Groups: A group of objects exhibit strong spatial relationships that interact with each other. 3. Relationships: This section illustrates the interactions or spatial relationships between various objects or groups. 4. Caption: Caption should be a simple and straightforward 2-4 sentence image caption.
Please include all the objects in the caption and refer to them in '()'. Create the caption as if you are directly observing the image. Do not mention the use of any source data. Do not use words like 'indicate', 'suggest', 'hint', 'likely', or 'possibly'.\\
\\
You can refer to the following examples as references. \\
\textit{In-context learning examples for Description Generator}\\
\\
Please provide a json format with Description based on the following object list. \\
\end{tcolorbox}
\end{center}

\begin{figure*}[t]
    \centering
    \includegraphics[width=0.99\linewidth]{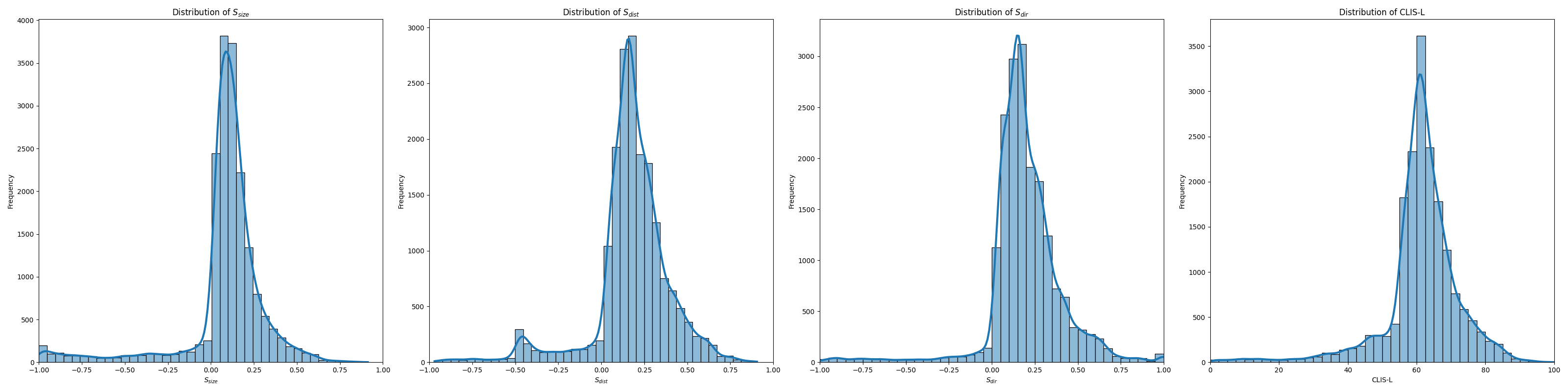}
    \caption{Score distributions of $S_{size}$, $S_{dist}$, $S_{dir}$, and CLIS-L.}
    \label{fig:CLIS-L_dist}
\end{figure*}

\begin{center}
\begin{tcolorbox}
\textbf{Prompt for Layout Generator}:\\
Task Description:\\
Your task is to generate a layout based on a detailed description.
The layout is a list of json with 'object' and 'bbox'. 'object' refers to the object name in the prompt provided, while 'bbox' is formulated as [x,y,w,h], where "x,y" denotes the top left coordinate of the bounding box. "w" denotes the width, and "h" denotes the height. The bounding boxes should not go beyond the image boundaries. The six values "x,y,w,h,x+w,y+h" are all larger than 0 and smaller than 1.\\
\\
You can refer to the following examples as references. \\
\textit{In-context learning examples for Layout Generator}\\
\\
Please provide a json format with Layout based on the following prompt.\\
\end{tcolorbox}
\end{center}

\subsection{Image Generator}
\label{app:image-generator}
We use InstanceDiffusion~\cite{wang2024instancediffusion} as the image generator. 
We follow the default settings of SDXL used in InstanceDiffusion to refine synthetic images for our image generator.
Regarding the number of synthetic images, we follow the approach used in StableRep~\cite{tian2024stablerep} and SynCLR~\cite{tian2023learning}, generating four images for each scene graph.

\subsection{Image Filter}
\label{app:image-filter}
For caption model $F_C$ in Eq.~\ref{equ:image-metric}, we adopt a pre-trained VLM, Qwen-VL~\cite{Qwen-VL}, which has strong perception abilities.
For $F_{sim}$ function in Eq.~\ref{equ:image-metric},
we prompt an LLM to assign a score based on text similarity. 
Unlike direct comparisons of text embeddings, LLMs can weigh different parts of the descriptions according to their significance.
For instance, a mismatch in categories results in a lower score than a mismatch in attributes.

\begin{center}
\begin{tcolorbox}
\textbf{Prompt for Caption Model}:\\
You are my assistant to evaluate the correspondence of the image to a given text prompt. \\
Briefly describe the image within 50 words. Focus on the objects in the image and their attributes (such as color, shape, texture), spatial layout, and action relationships.
\end{tcolorbox}
\end{center}

\begin{center}
\begin{tcolorbox}
\textbf{Prompt for Similarity Compution}:\\
You are an intelligent chatbot designed to evaluate the correctness of generative outputs for question-answer pairs.\\
Your task is to compare the predicted answer with the correct answer and determine if they match correctly based on the objects, and their actions, relationships. Here's how you can accomplish the task:\\
------\\
\#\#INSTRUCTIONS:\\
- Focus on the objects mentioned in the description and their actions and relationships when evaluating the meaningful match.\\
- Consider synonyms or paraphrases as valid matches.\\
- Evaluate the correctness of the prediction compared to the answer.\\ \\
Please Evaluate the following answer pair:\\ \\
Correct Answer: {answer}\\ 
Predicted Answer: {pred}\\ \\
Provide your evaluation in the JONSON format with the 'score' and 'explanation' key. The score is an integer value between 0 and 5, with 5 indicating the highest meaningful match. The explanation should be within 20 words.
\end{tcolorbox}
\end{center}

\section{Deployment Details on Downstream Tasks}
\label{app:deployment-details}

\subsection{Templates for Multi-modal Downstream Tasks}
\label{app:templates-for-multi-modal-downstream-tasks}
\begin{center}
\begin{tcolorbox}
\label{box:temp}
    \textbf{Localization}:\\
    Question:\\
    1. Where is the object described \{attribute\} located in the image in terms of the bounding box?\\
    2. What is the location of object described \{attribute\} in terms of the bounding box?\\
    3. Localize the object described \{attribute\} in terms of bounding box.\\
    4. Provide a bounding box for the object described \{attribute\}.\\
    5. Generate a bounding box for the object described \{attribute\}.\\
    6. Describe the object located at \{layout\}.\\
    7. Provide a caption for the object at \{layout\}.\\
    8. What is at location \{layout\} in image?\\
    Answer:\\
    1-5: It is located at \{layout\}.\\
    6-8: There is a \{attribute\}.\\

    \textbf{Attribute-binding}:\\
    Question:\\
    1. What is the color of \{obj\}?\\
    2. What color is the \{obj\}?\\
    3. What color do you think the \{obj\} is?\\
    4. Which color is the \{obj\}?\\
    5. What is the number of \{obj\}?\\
    6. What is the total count of \{obj\} in the image?\\
    Answer:\\
    1-4: \{color\}.\\
    5-6: \{number\}.\\

    \textbf{Relation}:\\
    Question:\\
    What is the relationship between the subject described \{attribute1\} and the object described \{attribute2\}?\\
    Answer:\\
    \{subject\} \{relation\} \{object\}.
\end{tcolorbox}
\end{center}

\noindent
We provide templates for constructing question-answer pairs for multi-modal downstream tasks. 
For perception, we design two types of tasks: localization and attribute-binding. 
Localization tasks necessitate that models pinpoint an object detailed in the instructions or, alternatively, describe an object situated at a specific location.
Attribute-binding tasks require models to identify precise attributes of an object within a given location or give a precise number of the target object. 
For reasoning, we craft relation reasoning tasks. These tasks require models to deduce the relationship between a specified subject and object based on the provided description.

\begin{figure*}[t]
    \centering
    \includegraphics[width=0.99\linewidth]{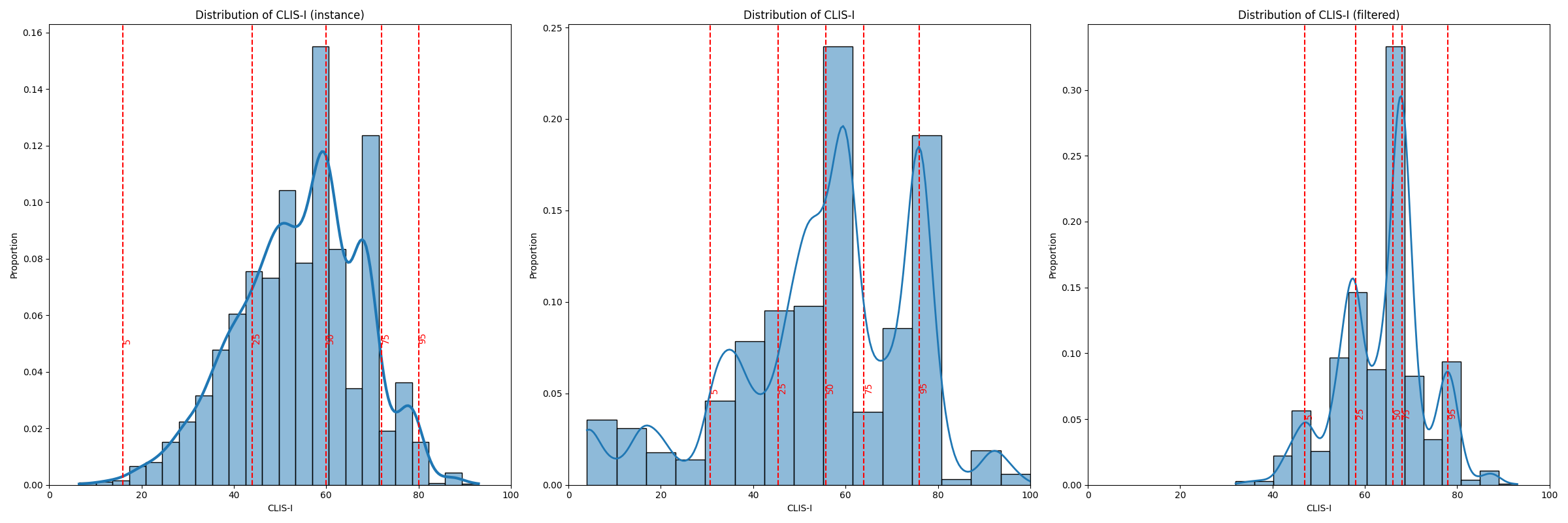}
    \caption{Score distribution of CLIS-I.}
    \label{fig:CLIS-I_dist}
\end{figure*}

\subsection{CLIS Settings}
\label{app:clis-settings}

\noindent
\textbf{CLIS-L Computation.}
We detail CLIS-L computation as follows:
\begin{enumerate}[label={\bf {{$\bullet$}}}, leftmargin=*, topsep=0.5ex, itemsep=-0.5ex, partopsep=0.75ex, parsep=0.75ex, partopsep=0pt, wide, labelindent=0pt]
    \item Penalty Function. To filter out noise data identified by a low score in any of the three metrics, $S_{size}$, $S_{dist}$, and $S_{dir}$, we introduce a penalty function $f$ and a score threshold $t$. The function $f$ is a linear transformation that maps scores below $t$ from $0$ to $t$ to a range of $-1$ to $t$.
    \item Weight. To balance the impact of the three metrics (size, distance, and direction) in Eq.~\ref{equ:clis-l}, we apply Z-score normalization to each. The distribution of scores across these three metrics and CLIS-L is shown in Fig.~\ref{fig:CLIS-L_dist}.
    \item Percentile Operation. We use percentile operation in Eq.~\ref{equ:clis-l}. 
    We first compare the percentile operation with the average operation. Given that multiple reasonable layouts can correspond to the same description, not all layouts from the data priors $P$ provide the necessary information for accurate assessment. For example, in the description 'a person holds an umbrella', it would be unreasonable to evaluate a synthetic layout where the umbrella is in the person's right hand using ground truth layouts from $P$ where the umbrella is in the left hand. To compare these two operations quantitatively, we conduct experiments. We construct a test set of 10K samples generated by ACP, each containing two objects in the scene graph. We first swap the layouts of the two objects to determine if CLIS-L can assign a higher score to the original layout. Additionally, assuming that good layouts can produce better images, we compare the images selected by the two different CLIS-L calculation methods. Specifically, we use these two methods to independently select the top 25\% highest-scoring data (2.5K examples) and calculate the FID score for the corresponding images. Table~\ref{tab:app-comp-max-avg} shows that percentile operation in CLIS-L outperforms average operation.
    We then compare the percentile operation with the max operation. Since we use LLMs to construct data priors $P$, it may cause some errors in $P$. Thus, percentile operation is more robust against similar errors in synthetic layouts.
\end{enumerate}

\noindent
\textbf{CLIS Distribution and Setting.}
For visual perception tasks, we emphasize image quality by applying a threshold to CLIS-I. Specifically, we set an instance-level threshold of $60$. Images in which all instances fall below this threshold are excluded from the training set. Fig.~\ref{fig:CLIS-I_dist} shows the original distribution of instance-level CLIS-I (left) and overall CLIS-I (middle), as well as the distribution of CLIS-I after filtering (right). The instance filtering ratio is approximately $50\%$, while the image filtering ratio is around $15\%$. 
For multi-model perception and reasoning tasks, we apply both a threshold for CLIS-I and an additional threshold of $50$ for CLIS-L to ensure the generated layouts are reasonable. The CLIS-I threshold remains consistent with that used in visual perception tasks.

\begin{table}[t]
    \begin{center}
    \caption{Comparison of Max operation and Average operation in CLIS-L.}
    \label{tab:app-comp-max-avg}
    \begin{tabular}{l|cc}
        \toprule[0.1em]
        Operation & Accuracy$\uparrow$ & FID$\downarrow$  \\
        \midrule
        Average & 58.8 & 61.4 \\
        Percentile & 61.0 & 55.5 \\
        \bottomrule[0.1em]
    \end{tabular}
    \end{center}
\end{table}

\section{Details of Experiments Setup}

\begin{figure}[t]
    \centering
    \includegraphics[width=\linewidth]{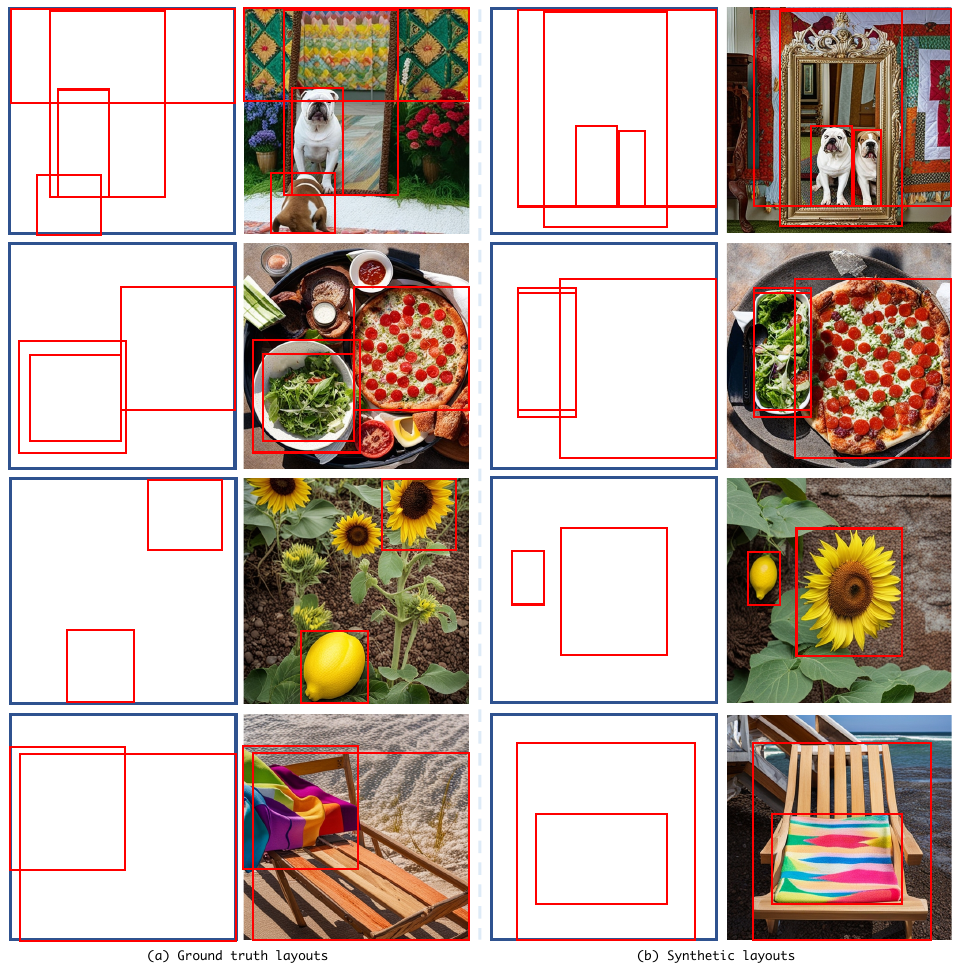}
    \caption{Comparison between ground truth layouts and synthetic layouts from our layout generator.}
    \label{fig:app-ablation-clis-l}
\end{figure}

\label{app:exp-setup}
\subsection{Baseline Settings}
Our specific baseline settings in experiments are as follows:
\begin{enumerate}[label={\bf {{$\bullet$}}}, leftmargin=*, topsep=0.5ex, itemsep=-0.5ex, partopsep=0.75ex, parsep=0.75ex, partopsep=0pt, wide, labelindent=0pt]
    \item Mask R-CNN baseline. We follow the same setup outlined in~\cite{gupta2019lvis}. Specifically, we adopt ResNet-50~\cite{he2016deep} with FPN~\cite{liu2020deep} backbone, using the standard 1$\times$ training schedule. 
    \item CenterNet2 baseline. We follow the setup outlined in~\cite{zhao2022x}. Specifically, we use two configurations: 1) ResNet-50 with a 1$\times$ training schedule, and 2) Swin-B with a 4$\times$ training schedule. We employ the AdamW optimizer and utilize repeat factor sampling with an oversample threshold of $10^{-3}$.
    \item Grounding-DINO baseline. We follow the setup outlined in ~\cite{zhao2024open}. Specially, we use the model pretrained on Objects365~\cite{shao2019objects365}, GoldG~\cite{kamath2021mdetr}, GRIT~\cite{peng2023kosmos}, and V3Det~\cite{wang2023v3det} with Swin-T~\cite{liu2021swin} as the backbone. The fine-tuning process uses the standard 1$\times$ training schedule. 
    We use AdamW~\cite{loshchilov2017decoupled} optimizer with a weight decay of 0.0001. The initial learning rate is 0.00005, dropped by 10$\times$ at the 8th and 11th epochs. 
    \item LLaVA-v1.5 baseline. We follow the setup outlined in ~\cite{liu2023improved}. We adopt a two-stage training process. For the LLM backbone, we adopt Vicuna-7B~\cite{vicuna2023}, Vicuna-13B, and LLama-3-8B~\cite{touvron2023llama}. 
    We use an AdamW optimizer with a weight decay of 0. Pre-training for 1 epoch with a 1e-3 learning rate and batch size of 32, and fine-tuning for 1 epoch with a 2e-5 learning rate and a batch size of 16. The warmup ratio of the learning rate is 0.03. 
    \item Stable Diffusion baseline. We use the v1-5 model weight from Huggingface~\cite{wolf2019huggingface}.
\end{enumerate}

\subsection{Training Settings}
We augment the original training set with synthetic examples to co-train downstream models, while annotations for rare categories are excluded in the open-vocabulary setting.

\begin{figure*}
    \centering
    \includegraphics[width=\linewidth]{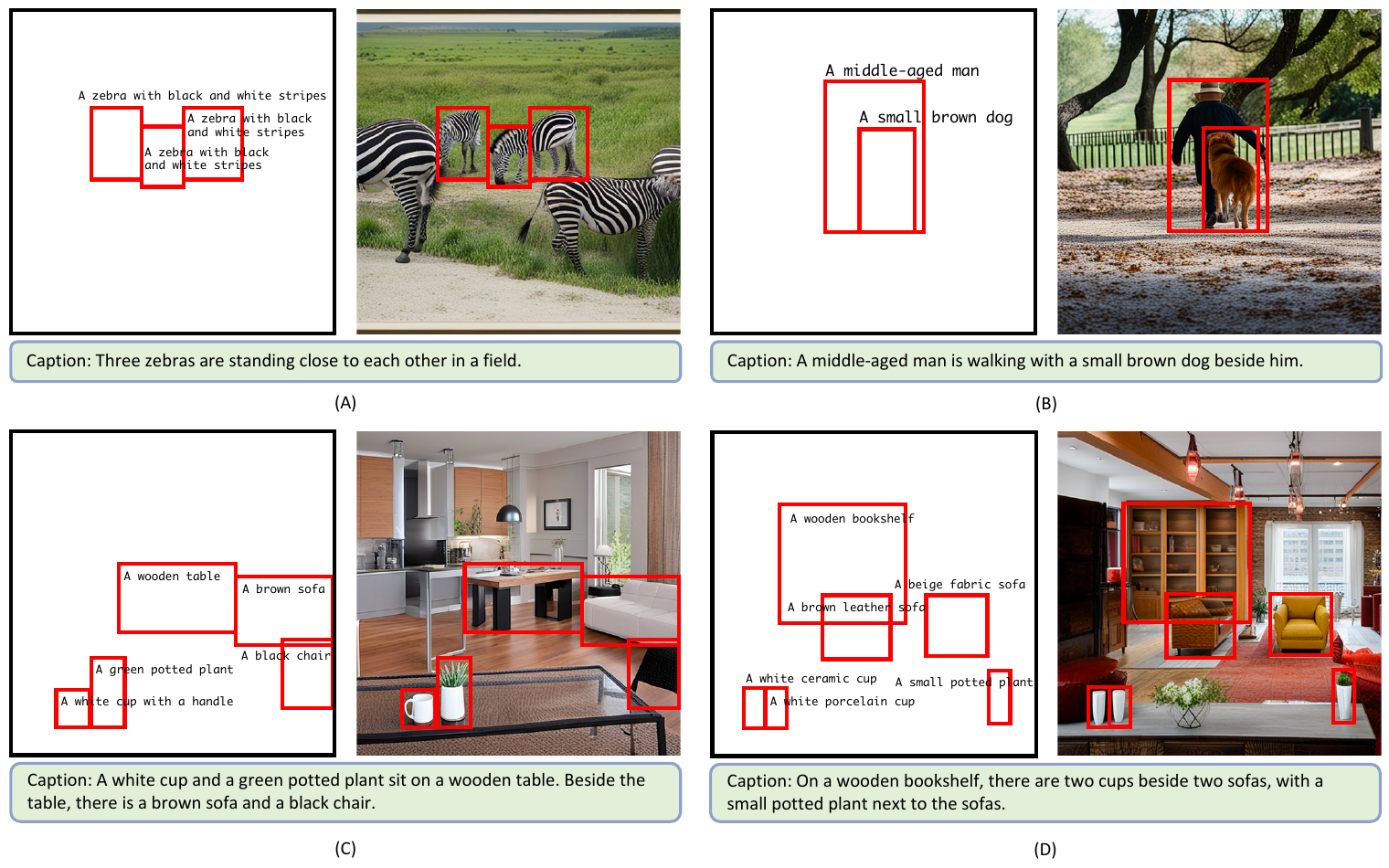}
    \caption{Error analysis of ACP. (A) Numerous objects and (B) overlapping objects for the image generator. (C)(D) complex object combinations for the scene graph generator. }
    \label{fig:app-error}
\end{figure*}

\subsection{Evaluation Protocols}
For generative metrics, FID is computed with the Inception V3~\cite{szegedy2016rethinking}. 
We adopt a pre-trained YOLOv8m following~\cite{wang2024instancediffusion} for YOLO score~\cite{li2021image} and report the standard average precision (AP), which is averaged at different IoU thresholds (from 0.5 to 0.95) across categories.

\subsection{Dataset Details}
MS-COCO is a common detection dataset containing 80 categories with 118K training images and 5K validation images. In the open-vocabulary setting~\cite{bansal2018zero}, MS-COCO can be divided into 48 base categories and 17 novel categories, excluding 15 categories without a synset in the WordNet hierarchy.

LVIS is a large vocabulary dataset with 1203 categories, featuring a long-tailed distribution of instances in each category. These categories can be divided into rare(337), common(461), and frequent(405) groups. LVIS training set contains 100K images, with an additional 20K images in the validation set.

The original instruction-following data mixture of LLaVA-1.5 is a total of 665K~\cite{liu2023improved}.

\section{Limitations and Future Work}
\label{app:limi}
Sampling of initial object combinations and evaluating layouts necessitates data priors $P$, which are resource-intensive to construct. Currently, $P$ is built from the COCO, LVIS, and Filter30K datasets. However, practical limitations in computational resources constrain our capacity to expand $P$, potentially impacting the accuracy of layout evaluation.
Generating high-quality samples through ACP is also computationally demanding, with a portion of synthetic samples to be filtered out. A future research direction involves developing computation-efficient methods to generate high-quality samples or devising strategies to learn from low-quality samples efficiently.

Additionally, simple experiments presented in Table~\ref{tab:app-comp-max-avg} and Fig.~\ref{fig:teaser}(a,b) indicate that better layouts contribute to improved image quality. Thus, another potential direction for future work is to use layout metrics to optimize computational resources in the generation process. We encourage more future studies focusing on the design of generation metrics. 

\section{Consistency with Human Preference}
\label{appx:consistency}
In Fig.~\ref{fig:con}, we present images generated from the same scene graph. The image quality consistently improves as the CLIS increases, confirming its alignment with human judgment.
To comprehensively evaluate consistency with human preferences, we additionally carry out a user study with 20 subjects. 
Each subject is shown 40 pairs of images, with each pair generated from the same scene graph with different CLIS scores.
The subjects are asked to evaluate the image pairs based on the following criteria:
\begin{enumerate}[label={\bf {{$\bullet$}}}, leftmargin=*, topsep=0.5ex, itemsep=-0.5ex, partopsep=0.75ex, parsep=0.75ex, partopsep=0pt, wide, labelindent=0pt]
\item Q1. choose the image that has the best \textbf{visual} quality.
\item Q2. choose the image that is better \textbf{aligned} with the annotation, including bounding boxes and text descriptions.
\end{enumerate}
A total of 1535 responses are collected. The results show that samples with higher CLIS get 66.1\% for Q1 and 94.7\% for Q2.
This indicates that higher CLIS aligns well with human judgments on visual quality and annotation alignment.

\section{Efficiency-Effectiveness Analysis}
\label{app:efficiency-effectiveness}

ACP demonstrates its efficiency: (1) Detectors efficiently utilize synthetic samples from ACP. For instance, X-Paste uses 100K synthetic images, double the size of ACP's synthetic dataset. (2) Synthetic data from ACP is richly annotated with detailed object attributes and relationships, making it readily applicable to various downstream tasks. (3) ACP significantly reduces the cost of data generation compared to manual collection and annotation, particularly for rare categories.

\section{Error Analysis}
\label{app:error-analysis}

Synthetic errors may arise in large-scale generation due to:
\begin{enumerate}[label={\bf {{$\bullet$}}}, leftmargin=*, topsep=0.5ex, itemsep=-0.5ex, partopsep=0.75ex, parsep=0.75ex, partopsep=0pt, wide, labelindent=0pt]
\item Scene Graph Generator. LLMs often struggle with rare or complex object combinations, leading to inaccurate layouts.
\item Image Generator. Diffusion models frequently fail when objects overlap or when rendering a large number of objects.
\end{enumerate}
As shown in Fig.~\ref{fig:app-error}, errors in (C) and (D) originate from the scene graph generator. When confronted with complex object combinations, LLMs may generate implausible layouts. For instance, in (C), the cup and plant should appear on the wooden table, and in (D), the two cups belong on the bookshelf.
Errors in (A) and (B) arise from the image generator. Diffusion models tend to struggle when handling (A) numerous objects or (B) overlapping objects.

\section{Visualization Results}
\label{app:vis}
\subsection{Ablation Study on Layout Generator}
\label{app:aba-lay}
We present visualizations of images generated using both ground truth layouts and synthetic layouts.
As shown in Fig.~\ref{fig:app-ablation-clis-l}, images generated with synthetic layouts exhibit comparable or even better to those generated with ground truth layouts.
Notably, ground truth layouts tend to overlap more, leading to low-quality results from diffusion models. 
Furthermore, synthetic layouts are more likely to be centered in the images, which helps reduce the occurrence of distracting objects in the generated images.

\subsection{Comparison with Other Metrics}
\begin{figure*}
    \centering
    \includegraphics[width=\linewidth]{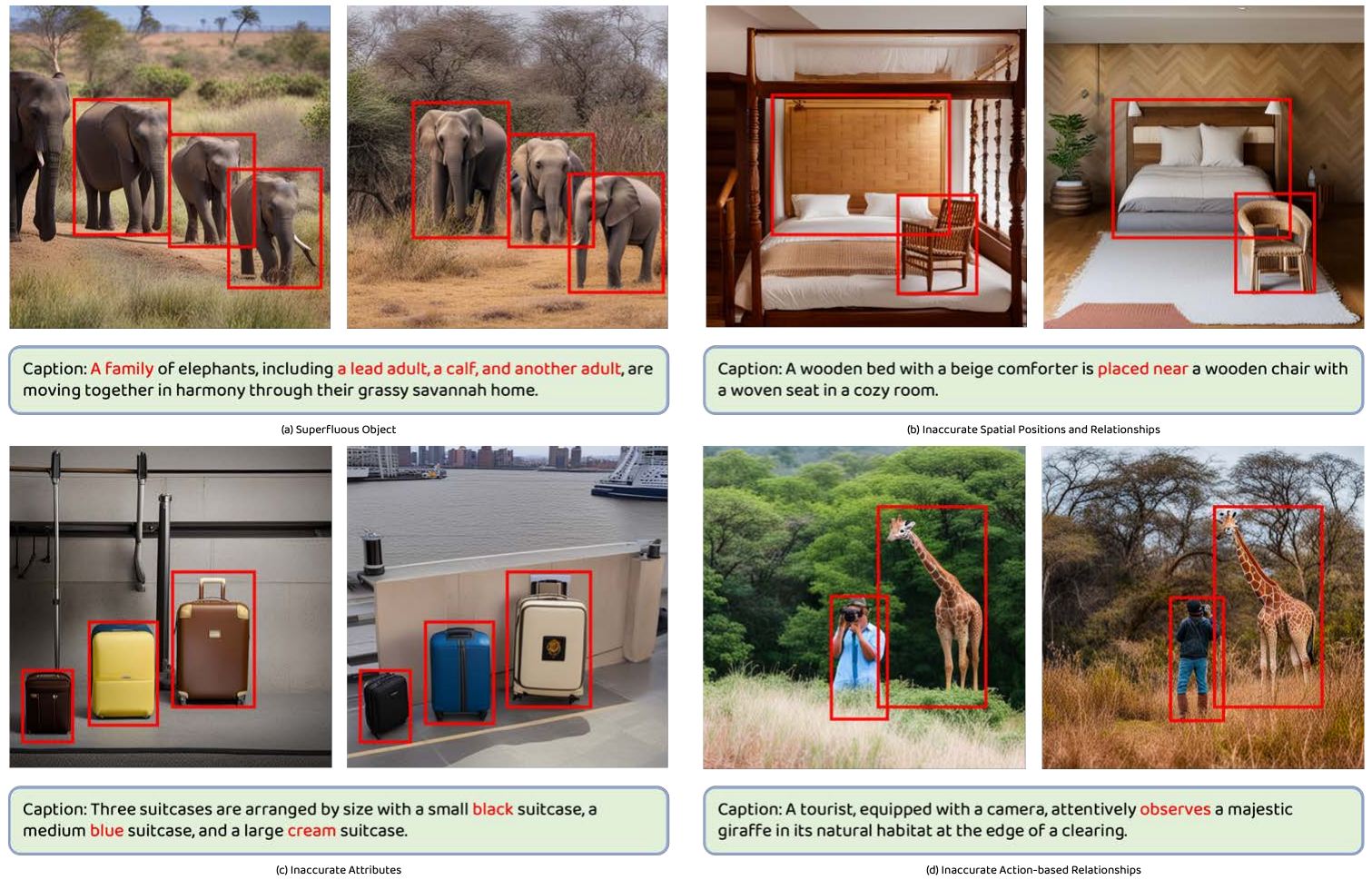}
    \caption{Comparison between CLIS-I and other prevalent metrics. Each pair of images is generated on the same scene graph, with CLIS-I favoring the right image in each pair. In (a) and (b), the CLIP score overlooks the extraneous elephant on the left and the inaccurate spatial arrangement between the chair and bed, respectively. For (c) and (d), the YOLO score fails to assess the detailed attributes or evaluate the semantic relationships between objects.}
    \label{fig:app-comp}
\end{figure*}

\noindent
\textbf{CLIS-I.}
We provide visual results comparing CLIS-I with other metrics. Using the same scene graph from the previous generator, we produce images evaluated with CLIS-I and other metrics, such as CLIP and YOLO scores. As illustrated in Fig~\ref{fig:app-comp}, CLIS-I demonstrates superior performance in both textual alignment and visual quality.

\begin{figure*}
    \centering
    \includegraphics[width=\linewidth]{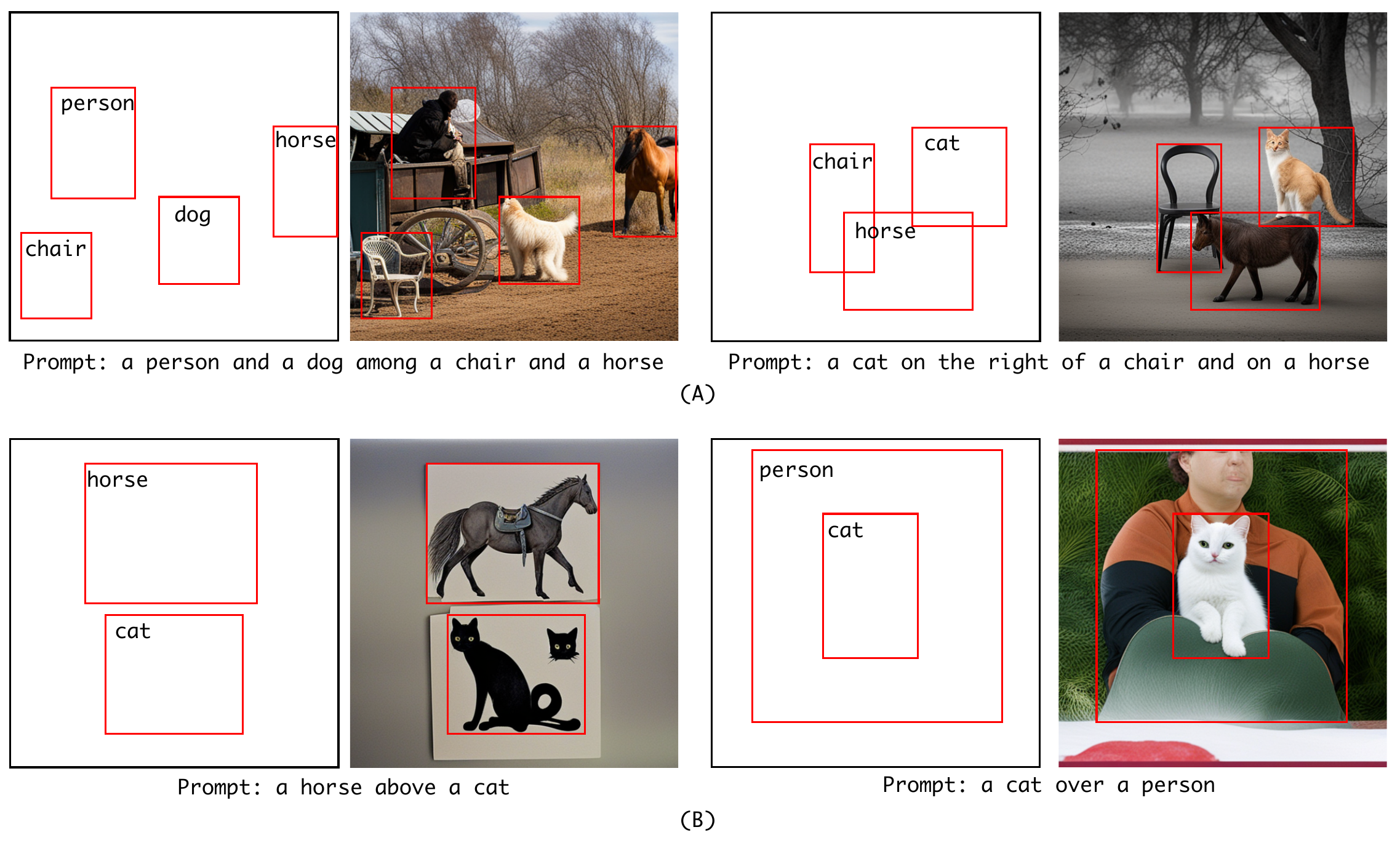}
    \caption{Comparison of CLIS-L and the HRS metric. (A) CLIS-L is consistent with the HRS metric in evaluating typical spatial relations. Both assign high scores to accurate spatial layouts. (B) CLIS-L provides additional filtering capability for problematic cases. For instance, the prompt 'A horse above a cat' is unreasonable in real-world scenarios. 'A cat over a person' is inaccurate as the cat should be positioned higher in the layout.}
    \label{fig:app-clis-l-comp}
\end{figure*}

\noindent
\textbf{CLIS-L.}
We further present visual comparisons of CLIS-L with the spatial detection-based HRS metric~\cite{bakr2023hrs}, similar to those used in T2I-CompBench~\cite{NEURIPS2023_f8ad010c}, which applies predefined rules to evaluate fix spatial relationships. 
To ensure that the relationships being evaluated are spatial and compatible with the HRS metric, we use the HRS spatial compositions benchmark~\cite{bakr2023hrs}. 
As shown in Fig~\ref{fig:app-clis-l-comp}(A), CLIS-L aligns with the HRS metric in evaluating typical spatial relationships. Both assign high scores to accurate spatial layouts. Fig.~\ref{fig:app-clis-l-comp}(B) highlights the advantage of CLIS-L, which assigns low scores to unrealistic or inaccurate spatial layouts, demonstrating its superiority in filtering suboptimal cases.
Notably, CLIS-L can also evaluate non-spatial layout relationships, further showcasing its versatility.

\subsection{Synthetic Training Examples}
\label{app:syn-samples}
\begin{figure*}
    \centering
    \includegraphics[width=\linewidth]{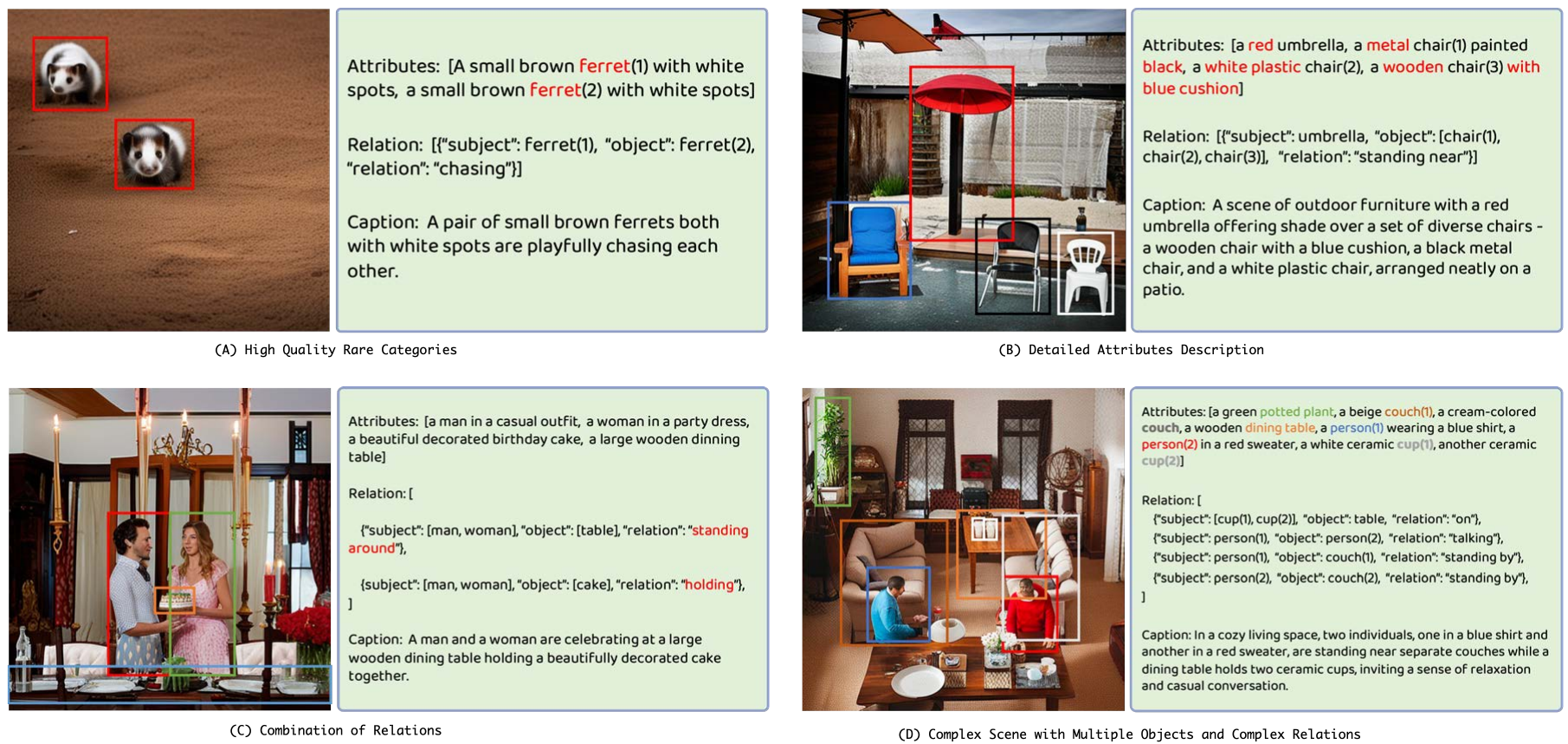}
    % \vspace{-3em}
    \caption{Synthetic training examples from ACP. In settings with imbalanced training data, such as long-tail scenarios, ACP can produce high-quality training examples for rare categories to mitigate this challenge. Additionally, ACP can generate diverse training samples with detailed attributes and relationships within complex scenes.}
    \label{fig:app-tra-long-tail}
\end{figure*}

\begin{figure*}
    \centering
    \vspace{-2em}
    \includegraphics[width=.73\linewidth]{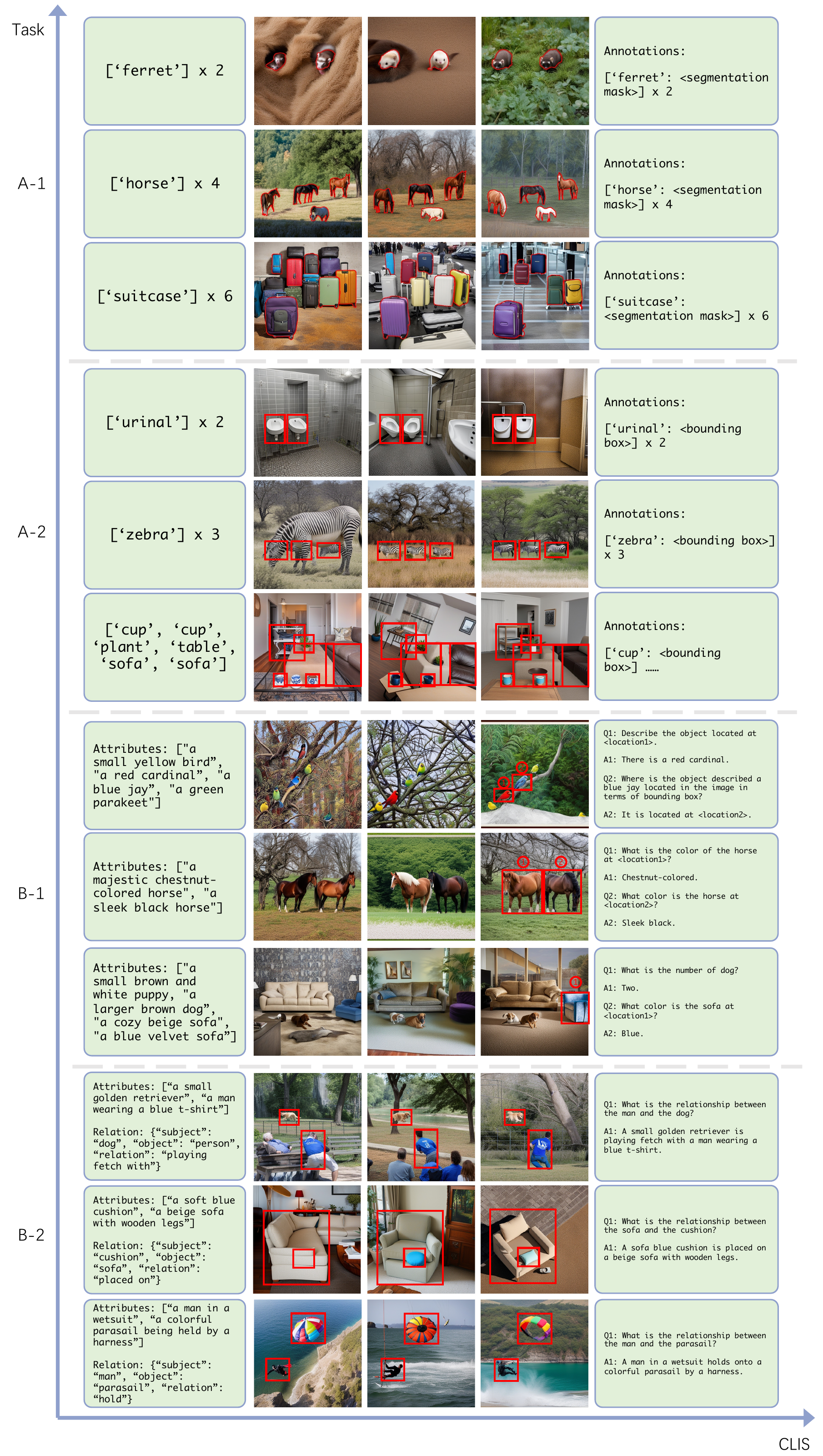}
    \vspace{-1em}
    \caption{\footnotesize Synthetic training samples of various tasks from ACP. Tasks A-1 and A-2 correspond to Segmentation and Detection, respectively. Tasks B-1 and B-2 pertain to multi-modal perception and reasoning. Given the same input or scene graph on the left, the CLIS of the synthetic training samples increases along the x-axis, with final annotations on the right.}
    \label{fig:app-tra-detection}
\end{figure*}

Additionally, we showcase visualizations of our synthetic training samples in Fig.~\ref{fig:app-tra-long-tail} and Fig.~\ref{fig:app-tra-detection}. By leveraging the extensive vocabulary of large generative models, we can produce high-quality training samples for rare categories.
These training samples are closely aligned with their respective scene graphs, capturing both detailed attribute descriptions and complex relationships between multiple objects effectively.

\end{document}